\documentclass[
    aps,
    prx,
    twocolumn,
    superscriptaddress,
    amsmath,
    amssymb,
    longbibliography]{revtex4-2}

\usepackage{comment}
\usepackage{enumitem}
\usepackage{graphicx}
\usepackage{hyperref}
\usepackage{cleveref}  
\usepackage{lipsum}
\usepackage{mathtools}
\usepackage{nicefrac}
\usepackage{xcolor}
\usepackage{cancel}
\usepackage{simpler-wick}
\usepackage{tikz}
\usepackage{notations}
\usepackage[normalem]{ulem}

\crefname{appendix}{appendix}{appendices}
\Crefname{appendix}{Appendix}{Appendices}
\crefname{figure}{Fig.}{Figs.}
\Crefname{figure}{Fig.}{Figs.}

\DeclareMathOperator{\var}{var}

\AddToHook{cmd/appendix/after}{%
  \crefalias{section}{appendix}%
  \crefalias{subsection}{appendix}%
  \crefalias{subsubsection}{appendix}%
}

\makeatletter
\let\oldparagraph\paragraph

\renewcommand{\paragraph}{%
  \vspace{3pt}%
  \@ifstar{\oldparagraph*}{\oldparagraph}%
}
\makeatother

\begin{document}

\title{Mesoscopic learning dynamics of extensive-width neural networks}

\author{Shu Zhou}
\affiliation{International School for Advanced
Studies (SISSA)\\ Via Bonomea 265, 34136 Trieste, Italy}

\author{Jean Barbier}
\affiliation{The Abdus Salam International Centre for Theoretical Physics (ICTP)\\Strada Costiera 11, 34151 Trieste, Italy}

\author{Sebastian Goldt}
\thanks{Correspondence: \{shuzhou, sgoldt\}@sissa.it, jbarbier@ictp.it.}
\affiliation{International School for Advanced
Studies (SISSA)\\ Via Bonomea 265, 34136 Trieste, Italy}

\begin{abstract}
  Understanding how stochastic gradient descent learns internal representations
  in wide neural networks trained on rich target functions is a central
  problem in machine learning. 
  We study this question in two-layer neural networks learning target functions
  represented by networks of the same architecture, with arbitrary nonlinearity, and with network width and
  input dimension growing proportionally. 
  In this regime, standard approaches for describing the learning dynamics
  require tracking a diverging number of overlaps, which measure correlations
  between learned and target features, leaving no obvious description in terms
  of a few order parameters.
  Here, we develop a mesoscopic description of learning by characterizing the
  \emph{distribution} of overlaps.
  In the early stages of learning, we find that a Gaussian description of
  overlap statistics reduces the learning dynamics to a few collective variables
  that track learning of the target function's linear and quadratic
  components.
  We then estimate the timescale of specialization, when Gaussian closure breaks
  down and individual learned features begin to strongly correlate with specific
  target features.
  We find that specialization in extensive-rank networks is accelerated via a
  \emph{quadratic shielding} mechanism, whereby learning the linear and
  quadratic components of the target function filters gradient fluctuations.
  We finally describe how the activation function determines the pattern of
  specialization and characterize the subsequent refinement of learned features.
  Together, these results show how collective learning gives way to
  individual feature discovery and clarify when
  learning in large neural networks admits a simple statistical description.
\end{abstract}

\maketitle

\section{Introduction}

The remarkable success of deep learning is inseparable from the ability of
stochastic gradient descent (SGD)~\citep{robbins1951stochastic} to extract
structure from high-dimensional data while navigating highly nonconvex loss
landscapes~\citep{lecun2002efficient, bottou2007tradeoffs}. Statistical physics
has provided a natural language for understanding such high-dimensional dynamics
since the seminal work of \citet{saad1995line, saad1995exact}, who analyzed the
generalization dynamics of two-layer neural networks in the limit of large input
dimension. The crux of their approach is to reduce the high-dimensional dynamics
of the parameters to a set of low-dimensional ordinary differential equations
(ODEs) describing a few macroscopic order parameters, or sufficient statistics,
which completely capture the dynamics of the generalization error. The practical
success of neural networks triggered renewed interest in this
framework~\citep{goldt2019dynamics}, which has since been made
rigorous~\citep{goldt2019dynamics,Veiga_ODE,arnaboldi2023high} and applied to
over-parameterization~\citep{goldt2019dynamics, richert2022soft}, continual and transfer learning~\citep{lee2021continual,
gerace2022probing, saglietti2022analytical, lee2024animals}, reinforcement
learning~\citep{patel2025rl}, the derivation of optimal learning
protocols~\citep{mori2025optimal, mignacco2026statistical},
and the impact of data structure on learning~\citep{goldt2020modeling,
refinetti2022dynamics, braun2025learning, braun2026fast}, among others.

The Saad and Solla approach requires tracking
a number of order parameters that scales quadratically with the number of
trainable neurons in the hidden layer. It is therefore limited to narrow
networks, for which the target is a single- or multi-index
model~\citep{benarous2021online, benarous2022high, damian2023smoothing}. A related
line of work analyzes homogenized SGD to derive deterministic
Volterra integral equations~\citep{paquette2022implicit}, and matrix-valued ODEs for
generalized linear and multi-index models~\citep{collins2024hitting},
always assuming a fixed number of learned and target
directions. In the
complementary limit of neural networks with infinitely wide hidden
layers, recent works have developed two different descriptions of the learning
dynamics. In the so-called ``lazy'' regime~\cite{chizat2019lazy}, the network
remains close to its initialization and behaves like a kernel method governed by
the neural tangent kernel (NTK)~\citep{jacot2018neural}. Since the kernel
remains constant in the infinite-width limit, this limit cannot capture
feature learning, a hallmark of neural
networks~\citep{hinton2012deep}. In the mean-field regime~\citep{mei2018mean,
chizat2018global, rotskoff2018neural, sirignano2020}, the hidden neurons are
interacting particles whose empirical distribution follows a nonlinear
partial differential equation (PDE). This description captures feature learning, but the
distribution lives in a space whose dimension grows with~$d$, which makes it hard to
analyze without extra structure.

\begin{figure*}[t!]
    \centering
\includegraphics[width=\textwidth]{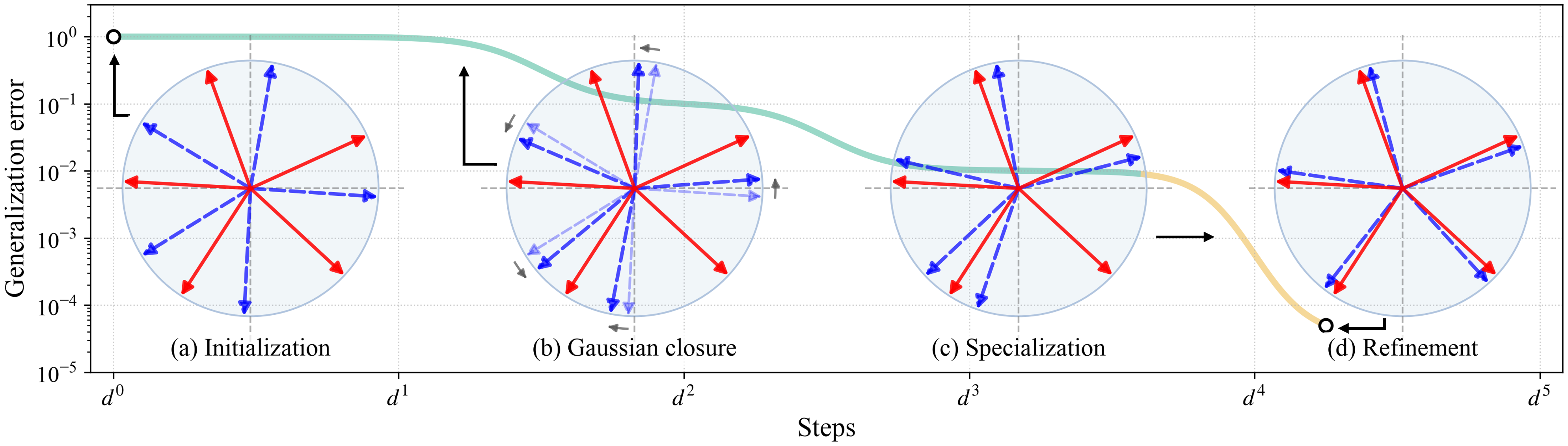}
    \caption{\label{fig:vis} \textbf{The stages of learning in an
    extensive-width neural network.} Schematic view of the learning regimes in
    the extensive-width teacher--student model. Red and blue arrows denote
    teacher and student hidden neurons, respectively, and the sphere represents
    the space of weight directions. The colored line schematically illustrates
    the decrease in generalization error during training. (a) At initialization,
    student weight vectors are isotropic and have only small random overlaps
    with the teacher's. (b) In the early stages of learning, the student learns
    a quadratic approximation of the teacher function by learning the teacher
    weights up to a rotation; individual teacher-student overlaps remain small
    $d^{-1/2}$. This ``Gaussian-closure'' regime is described by the simple
    closed equations derived in \cref{sec:gaussian-closure} and accounts for the
    first two stages of loss reduction, shown by the green curve. (c) At the
    specialization transition, a positive fraction of student neurons develop
    macroscopic correlations with  teacher neurons, although the matching may
    remain imperfect. Gaussian closure predictions fail, and the loss undergoes
    a third stage of reduction shown by the yellow curve; see
    \cref{sec:specialization}. The activation function determines the type of
    specialization; see \cref{sec:specialization}. (d) In the refinement phase,
    matched student neurons converge to their corresponding teacher directions,
    see \cref{sec:refinement}.}
\end{figure*}

In between these two extremes is the ``proportional'' or extensive-rank regime
where both input dimension and network width go to infinity at the same rate.
While closer to practice, it has proven hard to analyze. For the static
problem, where one seeks the Bayes-optimal error achievable from a number of 
samples $n$ scaling like the number of trainable parameters, recent work has located learning transitions
in extensive-width networks with quadratic~\citep{maillard2024bayes} and generic
activation functions~\citep{barbier2026statistical} using tools from statistical
physics and random matrix theory, following initial empirical work by
\citet{cui2023bayes}. While this line of work has already seen several
intriguing applications~\citep{erba2026nuclear, girardin2026width}, it does not
cover the learning dynamics of a network trained with SGD. The learning dynamics
of a wide student with quadratic activation function on a single-index target
was studied by \citet{sarao2020optimization} for phase retrieval and more
recently by \citet{martin2024impact, martin2026high}; the latter analyze
recovery of the full weight matrix, and do not explicitly study neural alignment
between student and teacher. In any case, we will see that the higher-order
terms play a crucial role in the dynamics of general networks.
\citet{ren2025emergence, benarous2025learning} considered feature learning in
growing-rank regimes. Conceptually close to our work, \citet{berthier2025learning} use an elegant
perturbative approach to derive learning timescales for a wide two-layer neural
network trained on a single-neuron teacher when the second layer is trained
faster than the first; \citet{gerbelot2026singular} later made the analysis of
the early-stage learning rigorous. Here, we focus on the case where both teacher
and student are extensive and there is no timescale separation.
We discuss further related work in \cref{sec:related-work}.
\vspace*{-2em}
\subsection{Contributions}

Here, we revisit the Saad--Solla approach for two-layer neural networks
with arbitrary nonlinear activation in the extensive-width limit $k,d\to\infty$ at fixed $k/d$. We focus on the teacher--student setup, where inputs $\bx$ are $d$-dimensional Gaussian
vectors with mean zero and identity covariance, and labels~$y$ are provided by a
fixed teacher network with random weights. Teacher and student have equal width
and fixed all-ones readouts. 
Expanding
the activation function in Hermite polynomials reveals a hierarchy of
tensor-contracted order parameters akin to \citet{barbier2026statistical}. The
dynamics of these order parameters delineate the stages of learning summarized in
\cref{fig:vis}: isotropic initialization (a), weak collective alignment, called ``Gaussian-closure regime''  (b),
specialization (c), and refinement (d).

Our main finding is that the dynamics close on two order parameters before
specialization, but not after. In the early stages of learning, the student only
learns the first two Hermite modes of the teacher function, and its dynamics reduce
to an autonomous low-dimensional system of ODEs under a ``Gaussian closure''
assumption on the correlations between teacher and student neurons. Specialization,
on the other hand, is driven by higher-order Hermite modes; although we find no
closed low-dimensional system in this regime, we can still predict analytically
when the network specializes. 

In the following, we define our setting in
\cref{Sec:Framework}. Starting from two observations about the learning dynamics in \cref{sec:observations}, 
our main contributions are:

\begin{enumerate}[label=(\roman*)]
\item \textit{Gaussian closure in the pre-specialization regime.} We derive a
  low-dimensional autonomous flow that describes the learning of the teacher's
  linear and quadratic components and predicts the generalization loss before
  specialization. The flow closes on a small number of interpretable order
  parameters under a \emph{Gaussian-closure} assumption: finite collections of
  teacher--student and off-diagonal student--student overlaps are treated as
  jointly Gaussian, see (\cref{sec:gaussian-closure}).
\item \textit{The onset of specialization and quadratic shielding.} We estimate
  the time at which a linear fraction of student neurons develops strong
  correlations with individual teacher neurons (\cref{fig:vis}(c)). The
  timescale is governed by the \emph{specialization exponent} $r=r(\sigma)$, the
  smallest degree strictly greater than two at which the Hermite expansion of
  $\sigma$ has a nonzero coefficient, and reads 
    \begin{equation*}
    \tau_{\rm spec}
    =\Theta\Big(\frac{k^r}{(\log k)^{r-2}}\Big),
    \end{equation*}
  We also identify a \emph{quadratic shielding} mechanism whereby having learned
  the low-order components of the teacher filters gradient fluctuations
  accelerates specialization by an activation-dependent constant, and discuss
  the different patterns of specialization, see \cref{sec:specialization}.
\item \textit{The refinement phase.} When learning arrives at a stage with a
  fixed one-to-one assignment of student to teacher neurons, we derive the
  dynamics of the matched overlaps and their fixed points, see 
  \cref{sec:refinement}.
\end{enumerate}

\vspace*{-2em}
\subsection{Further related work}%
\label{sec:related-work}

\paragraph*{Plateaus and specialization in narrow networks.}

\citet{saad1995line, saad1995exact} analyzed specialization in
committee machines and found that the generalization error passes through an
extended plateau before its final decay. They traced this plateau to an unstable
fixed point of the order-parameter ODEs at which all student neurons
have the same finite overlap with all teacher neurons. Escaping the plateau
requires breaking this permutation symmetry, which happens when small initial
asymmetries in the overlaps grow exponentially along the unstable direction of
the symmetric fixed point. A more precise analysis of these
transients~\citep{biehl1995learning, riegler1995line, biehl1996transient} in
showed that the plateau length diverges with the inverse of the logarithm of the initial asymmetry
in rescaled ``time'' $t = n / d$, where $n$ is the number of steps. More recently, the plateau has been revisited
for modern activation functions~\citep{yoshida2019data,
straat2019line,tian2020student, oostwal2021hidden}. All of these works require a
finite number of hidden neurons, whereas our focus is on the extensive-rank
case.

\paragraph*{Identifying learning timescales via loss invariants for finite-rank teachers.}

One of our key results is the timescale at which neurons start specializing. A large body of recent work has sought to express the timescales of learning
with online SGD, and hence its sample complexity, in terms of a few invariants
of the population loss. \citet{benarous2021online} introduced the \emph{information
exponent} $k^\star$ of a single teacher neuron, i.e., the degree of the first
non-vanishing Hermite coefficient of its activation function, and showed that
online SGD requires $\Theta(d^{\max(1,k^\star-1)})$ samples (up to logarithmic
factors) to escape the uninformative region around a random initialization. The
same exponent governs learning with shallow neural
networks~\citep{bietti2022learning} and controls the trade-off between time and
sample complexity of online SGD with large batches~\citep{arnaboldi2024online}.
For multi-index targets, the \emph{leap exponent} of \citet{abbe2022merged,
abbe2023sgd} measures the largest jump in degree along the hierarchical
``staircase'' in which the relevant directions are learned one after the
other~\citep{abbe2021staircase}, and sets the duration of the resulting
saddle-to-saddle dynamics. \citet{bietti2023learning} characterize
similar saddle-to-saddle dynamics and Hermite-dependent timescales under a
two-timescale learning procedure for low-rank target functions. These
invariants also depend on the input
distribution: correlations between latent variables lower the effective
information exponent and accelerate learning~\citep{bardone2024sliding}, and so
does a spiked input covariance~\citep{mousavi2023gradient}.

\paragraph*{Sequential learning in the ``giant-steps'' framework.}

For extensive-rank students trained on single-
or multi-index teachers when the number of samples $n$ scales as a power of
$d$, \citet{ba2022high} showed that one step of gradient descent on the first
layer creates alignment with the teacher's linear component and, with a
sufficiently large learning rate
and subsequent readout fitting, can yield nonlinear prediction beyond
random-feature and kernel baselines. Related one-step representation-learning
guarantees are established by \citet{damian2022representations}.
\citet{dandi2024two, cui2024asymptotics} extended this analysis to several
steps of gradient descent, and found that target directions are learned
sequentially if they satisfy the staircase property. We also find that the
student learns different parts of the teacher function sequentially, but we
follow the continuous online evolution for a teacher with extensively many
features.

\paragraph*{Dynamical mean-field theory of feature learning.}

The key object in the Saad--Solla approach is the matrix of teacher--student overlaps,
which becomes a large matrix in the extensive regime that we consider. A
complementary description of the learning dynamics is provided by the dynamical
mean-field theory (DMFT) of wide networks; see \citet{bordelon2026disordered}
for a recent review, and see \citet{mignacco2020role, mignacco2020dynamical} for
earlier work using DMFT for small neural networks, which was made rigorous by 
\citet{celentano2021high,gerbelot2024rigorous}. These works typically
consider infinite width at fixed input dimension and fixed training set; the
dynamical order parameters are then the feature and gradient kernels evaluated
on all pairs of training samples and training times, with the hidden-neuron
index averaged out~\citep{bordelon2022self}, with $1/\sqrt{\text{width}}$
fluctuations around the deterministic limit characterized
in~\citet{bordelon2023dynamics}.  This description captures feature learning in
the mean-field parametrization, but the number of order parameters grows with
the number of samples and of time steps rather than with the input dimension,
and no extensive teacher is present. \citet{montanari2026dynamical} derived the
DMFT of gradient flow for a two-layer network with generic activation in the
limit $n,d\to\infty$ at fixed $n/d$, followed by a large-width limit, but for a
teacher with at most a few neurons. Finally, \emph{adaptive-kernel} theories of feature learning consider Bayesian or
Langevin-trained networks at equilibrium, rather than by SGD~\citep{li2021statistical,pacelli2023statistical,seroussi2023separation, fischer2024critical,rubin2025kernels}.

\section{Teacher--student setup and order parameter dynamics}
\label{Sec:Framework}

\subsection{Extensive-width teacher--student setup}

\paragraph*{Task and architecture.} We consider a supervised learning task where inputs $\bx_t \in \mathbb{R}^d$ are sampled i.i.d. for $t=1, \dots, \tau$ from the multivariate normal distribution with zero mean and identity covariance~$\mathsf{N}(0, I_d)$. The corresponding responses~$(y^0_t)$ are generated by a two-layer teacher network with $k$ hidden neurons as
\begin{equation}
    \label{eq:teacherModel}
    y^0_t\coloneq y^0(\bx_t)\coloneq  \frac{\mathbf{1}^\intercal}{\sqrt{k}} \sigma\Big(\frac{\bW^0 \bx_t}{\sqrt{d}}\Big)
\end{equation}
where $\bW^0 \in \mathbb{R}^{k \times d}$ is the fixed first-layer weight matrix with i.i.d.\ standard Gaussian entries; the labels are noiseless. The activation function $\sigma:\mathbb{R}\to\mathbb{R}$ is applied element-wise.

The student network possesses the same architecture and produces the prediction
\begin{align}
    \label{eq:studentModel}
    y_t\coloneq y(\bx_t)\coloneq  \frac{\mathbf{1}^\intercal}{\sqrt{k}} \sigma\Big(\frac{\bW \bx_t}{\sqrt{d}}\Big).
\end{align}
When the network label is needed, we write $y\equiv y^1$ and $\bW\equiv\bW^1$. Networks with fixed second-layer
weights, also called committee machines, are a standard model of neural
networks~\cite{schwarze1993learning, saad1995exact,
engel2001statistical, aubin2018committee, goldt2019dynamics}. This setting
isolates feature learning in the first layer. Trainable readouts would require
additional order parameters; we leave them, and deeper networks, to future
work.

\paragraph*{Training.} We train the student by one-pass (``online'') stochastic
gradient descent (SGD): at each step $t$, we compute the gradient of the square
loss on a single fresh sample $\bx_t\sim \mathsf{N}(0, I_d)$ and its teacher label
$y^0_t$, \cref{eq:teacherModel}. The training time therefore equals the number of
training samples. This online setting is common in theory~\cite{saad1995line,
saad1995exact}; while only an approximation for, e.g., image classification, where
examples are reused, it is close to current practice for large language models,
where most inputs are used only once or twice~\citep{brown2020language,
touvron2023llama}. We initialize $\bW$ with i.i.d.\ standard Gaussian entries and
update it as
\begin{align}
    \label{eq:sgd} 
    \bg(t) &= -\nabla_\bW \big(y^0_t - y(\bx_t,\bW)\big)^2\big|_{\bW=\bW(t)} \nonumber\\
    \bW(t+1) &= \sqrt{kd}\frac{\bW(t) + \eta\, \bg(t)}{\|\bW(t) + \eta\, \bg(t)\|_{\rm F}},    
\end{align}
where $\bg(t)\in \mathbb{R}^{k\times d}$ is the gradient matrix and
$\|\,\cdot\,\|_{\rm F}$ is the Frobenius norm. We hence impose a
global spherical constraint on the student weight matrix, $\|\bW(t)\|_{\rm
F}^2=kd$. This constraint fixes the average squared norm of the hidden-neuron
weights, but does not constrain each hidden neuron separately. In the regimes
considered below, we observe that the individual hidden-neuron norms remain
concentrated around this average. 


\paragraph*{Scaling regime.} We consider the following high-dimensional, extensive-width limit
\begin{align*}
    d,k\to \infty \quad \text{with}\quad \frac{k}{d}\to  \gamma,\quad  \gamma \in (0,\infty).
\end{align*}
Since $\gamma$ is fixed, powers of $k$ and $d$ are interchangeable in asymptotic
statements; we write size exponents in powers of $k$. 
We take the learning rate to vanish in the extensive-width limit,
$\eta\to0$ as $d\to\infty$; the required rate depends on
the dynamical regime. This places the dynamics in the
gradient-flow, or perfect-learning, regime identified by
\citet{Veiga_ODE} and \citet{arnaboldi2023high}, in which the contributions of the
effective SGD noise vanish asymptotically; we return to this point in \cref{sec:master_eq}.


\subsection{Order parameters}

The activation function $\sigma$ can be decomposed in the orthogonal basis of Hermite polynomials $({\rm He}_{\ell}(x))_{\ell=0}^\infty$ defined as ${\rm He}_{\ell}(x)\coloneq \frac{d^\ell}{ds^\ell}\exp(sx-s^2/2)|_{s=0}$:
\begin{align}\label{eq:HermiteDecomp}
    \sigma(x)&=\sum_{\ell\ge 0} \frac{\mu_\ell}{\ell!} {\rm He}_{\ell}(x),
\end{align}
where $\mu_\ell=\mu_\ell(\sigma)\coloneq \mathbb E_{z\sim\mathsf N(0,1)}[\sigma(z)\,{\rm He}_\ell(z)]$ are the so-called Hermite coefficients of $\sigma$.

With this expansion, the student decomposes as
\begin{align*}
y &= \!\sqrt k\mu_0+\frac{\mu_1}{\sqrt{d}}\frac{\mathbf{1}^\intercal\bW}{\sqrt{k}}\bx\! +\! \frac{\mu_2}{2d}\operatorname{Tr}\Big(\frac{\bW^\intercal\bW}{\sqrt{k}}(\bx^{\otimes 2} - I_d )\Big) \!+ \!\cdots
\end{align*}
Note that the constant term in the expansion drops out of the loss, and hence the dynamics, since teacher and student share the same activation function. The $\cdots$ contains terms made of tensors of all orders constructed from $\bW$, contracted with input rank-one tensors $(\bx^{\otimes\ell})_\ell$. In each such term, at least one tensor is of order $\ell\ge 3$. Learning the network is thus a ``tensor-sensing'' problem: the labels depend on the tensors $\mathbf{1}^{\intercal}\bW^0\in \mathbb{R}^d,\ \bW^{0\intercal}\bW^0\in \mathbb{R}^{d\times d},\ldots$, all built from the same weights $\bW^0$. The first-order and second-order terms are called the ``linear component'' and ``quadratic component'', respectively.

Building on the framework of \citet{saad1995exact, saad1995line}, we
introduce the teacher--student
overlap matrices describing the system at microscopic neuron-wise resolution:
\begin{equation*} 
    \Omega_{ij}^{ab}\coloneq \frac{1}{d}\sum_{\alpha=1}^{d} W_{i\alpha}^{a}W_{j\alpha}^{b},
\end{equation*}
where $a,b\in\{0,1\}$ label the teacher ($a=0$) and student ($a=1$) networks,
with $\bW^1\equiv\bW$, and $i,j \in [k]$. We denote by $\mathbf w_i^a\in\mathbb R^d$
the $i$th row of $\bW^a$, so that $\Omega^{ab}_{ij}=\mathbf w_i^a\cdot\mathbf w_j^b/d$ and
$\bOmega^{01}=(\bOmega^{10})^\intercal$. The crucial difference from
\citet{saad1995exact, saad1995line} is that the overlap matrices are now
\emph{extensive}: their number of entries diverges. 

To handle the extensive overlap matrices, we describe them at the \emph{mesoscopic resolution}, i.e., we will track their distributions via their (rescaled) empirical moments: 
\begin{equation*}
    \begin{aligned}
        Q_p^{ab}&\coloneq \frac1k\sum_{i,j=1}^{k}(\Omega_{ij}^{ab})^p
        =\frac1{kd^{p}}\sum_{i,j=1}^{k}
        \prod_{m=1}^{p}\sum_{\alpha_m=1}^{d}W_{i\alpha_m}^{a}W_{j\alpha_m}^{b} . 
    \end{aligned}
\end{equation*}
This strategy follows the recent
mean-field theory of the statics of extensive-width networks by
\citet{barbier2026statistical}. The parameter $p$ indexes the degree of the ``overlap moment'' (or simply overlap). 
These overlaps describe how well the student learns each Hermite component of the teacher and separately capture its contribution to the loss. Indeed, the population loss admits the following exact decomposition~\cite{barbier2026statistical}:
\begin{equation}
    \label{eq:loss_decomp}
    \begin{aligned}
        \mathcal{L} \coloneq  \mathbb{E}(y(\bx)\!-\!y^0)^2
         &\! = \!      \sum_{\ell=0}^{\infty}\frac{\mu_\ell^2}{\ell!}\,
        \big(
        Q_\ell^{00}-2Q_\ell^{01}+Q_\ell^{11}
        \big). 
    \end{aligned}
\end{equation}
with $(\bx,y^0)$ a fresh sample generated via the teacher as in \cref{eq:teacherModel} and $y(\bx)$ the student prediction. 

To measure alignment between individual neurons, we also introduce the permutation-fixed order parameters, 
\begin{equation}
Q_{W_\ell}^{ab}
\coloneq 
\frac{1}{k d^\ell}
\max_{\bP}
\operatorname{Tr}
\Big[
\big(
\bW^a\bW^{b\intercal}\bP
\big)^{\odot\ell}
\Big],
\label{eq:Q_w}
\end{equation}
where the maximum is over $k\times k$ permutation matrices $\bP$, and $\odot\ell$ denotes the element-wise (Hadamard) $\ell$th power. For teacher--student overlaps, the maximization selects the assignment with the largest sum of powered overlaps, making the observable independent of the neuron labels in either network. The cases $\ell=1,2$ measure signed alignment and alignment in magnitude, respectively.\\

\subsection{Master equations for the order parameters} \label{sec:master_eq}

We assume that, in the extensive-width limit, the order parameters concentrate
on deterministic values throughout training, i.e., their stochastic SGD dynamics
are well approximated by their average, as established rigorously for two-layer
networks with few hidden neurons~\citep{goldt2019dynamics,Veiga_ODE,arnaboldi2023high}. 
We therefore omit sample averages in what follows and identify $Q_\ell^{ab}$
with these limiting observables. For every integer $p\geq 1$, the dynamics
of these order parameters during training with online SGD obey a
set of ODEs that can be obtained by substituting the Hermite
expansion of \cref{eq:HermiteDecomp} into the equations of
\citet{saad1995exact}; with our scaling of learning rate, these ``master equations'' read
\begin{widetext}
\begin{equation}
\begin{aligned}
Q_p^{01}(t+1)
= Q_p^{01}(t)
&+\frac{2p\eta}{kd}
\sum_{\ell=0}^{\infty}\frac{\mu_{\ell+1}^2}{\ell!}
\Big[
\frac{1}{k}\operatorname{Tr}\!\Big(
(\boldsymbol{\Omega}^{01})^{\odot(p-1)\intercal}
\big[
\bOmega^{00}(\bOmega^{01})^{\odot\ell}
-\bOmega^{01}(\bOmega^{11})^{\odot\ell}
\big]
\Big)
-Q_p^{01}(Q_{\ell+1}^{01}-Q_{\ell+1}^{11})
\Big]\\
&+\frac{2p\eta}{kd}
\sum_{\ell=0}^{\infty}\frac{\mu_\ell\mu_{\ell+2}}{\ell!}
\Big[
\frac{1}{k}\mathbf1^\intercal
(\bOmega^{01})^{\odot p}
\big[
(\bOmega^{01})^{\odot\ell}
-(\bOmega^{11})^{\odot\ell}
\big]^\intercal\mathbf1
-(Q_\ell^{01}-Q_\ell^{11})Q_p^{01}
\Big]
+O\Big(\frac{\eta^2}{d^2}\Big),
\\[1ex]
Q_p^{11}(t+1)
= Q_p^{11}(t)
&+\frac{4p\eta}{kd}
\sum_{\ell=0}^{\infty}\frac{\mu_{\ell+1}^2}{\ell!}
\Big[
\frac{1}{k}\operatorname{Tr}\!\Big(
(\bOmega^{11})^{\odot(p-1)\intercal}
\big[
\bOmega^{10}(\bOmega^{01})^{\odot\ell}
-\bOmega^{11}(\bOmega^{11})^{\odot\ell}
\big]
\Big)
-Q_p^{11}(Q_{\ell+1}^{01}-Q_{\ell+1}^{11})
\Big]\\
&+\frac{4p\eta}{kd}
\sum_{\ell=0}^{\infty}\frac{\mu_\ell\mu_{\ell+2}}{\ell!}
\Big[
\frac{1}{k}\mathbf1^\intercal
(\bOmega^{11})^{\odot p}
\big[
(\bOmega^{01})^{\odot\ell}
-(\bOmega^{11})^{\odot\ell}
\big]^\intercal\mathbf1
-(Q_\ell^{01}-Q_\ell^{11})Q_p^{11}
\Big]
+O\Big(\frac{\eta^2}{d^2}\Big).
\end{aligned}
\label{eq:Q_p}
\end{equation}
\end{widetext}
A detailed derivation is provided in \cref{Sec:exp_triple} for $p=1$ and extended in \cref{Sec:Overlap_Flow} to the general case $p>1$. Throughout, we omit the time index $t$ of $Q$
and $\bOmega$ in the driving terms. The master equations couple all $Q_p$ because
the gradients contain contractions of $\bOmega$.

The master equations can also be extended to include terms in $\eta^2$, which
depend on the second moments of the sample gradient. Here, we retain only the
terms linear in~$\eta$, consistently with the gradient-flow regime discussed by
\citet{Veiga_ODE, arnaboldi2023high}. In our normalization, the omitted terms
are $O(\eta^2/d^2)$ per update and therefore contribute only $O(\eta)$ through
the Gaussian closure regime, i.e.\ over a timescale $t=O(d^2/\eta)$. We thus
neglect them in the Gaussian-closure analysis, taking $\eta=o(1)$ as
$d\to\infty$; for example, $\eta=1/\log d$ suffices at this level of scaling.
This truncation need not remain valid on the longer specialization timescale. We
therefore retain the relevant quadratic terms when analyzing specialization in
\cref{sec:specialization}, where they determine the admissible learning rate.

\section{Results: the mesoscopic dynamics of learning}
\label{Sec:Results}

The master equations for the order parameters $Q_p$ are the starting point of
our analysis. Two numerical observations guide it: the $Q_p$ evolve on separated timescales, and the overlaps $\Omega_{ij}$ are Gaussian early in
training.

\subsection{Sequential learning and overlap statistics}%
\label{sec:observations}

\begin{figure}[t!]
    \centering\includegraphics[width=1\linewidth]{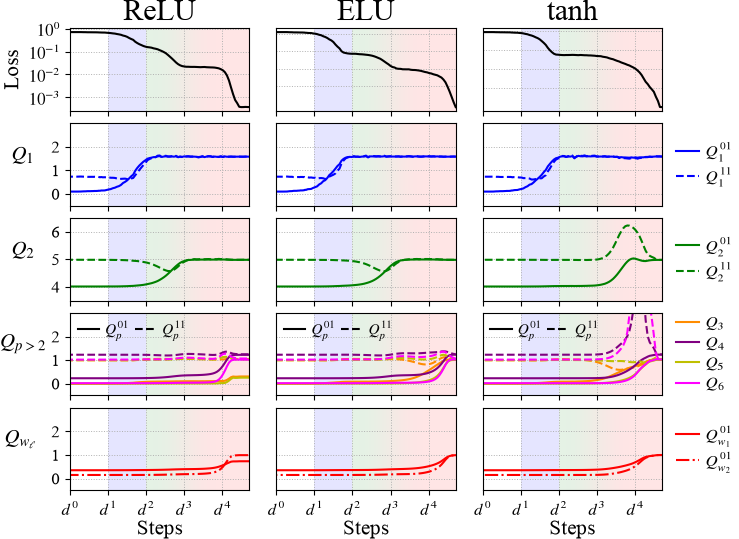}
    \caption{\label{fig:dynamics_closure_decomp} \textbf{Sequential learning of order
    parameters.} For the three different activation
    functions ReLU, ELU, and tanh, we show the test loss (top row) and the
    evolution of the order parameters $Q_p$ and $Q_{W_\ell}$ during training. The
    shaded regions indicate the successive regimes dominated by the dynamics of
    $Q_1$, $Q_2$, and higher-order modes $Q_{p\geq3}$, while the bottom row
    tracks a permutation-fixed student--teacher overlap. \emph{Parameters}:
    Network with width $k=200$, input dimension $d=50$.}
\end{figure}

\Cref{fig:dynamics_closure_decomp} shows the test loss of extensive-width networks
with various activations during training, together with the order
parameters $Q_1, Q_2$ and $Q_{p \ge 3}$ from the same simulations. Although all
order parameters in \cref{eq:Q_p} are coupled, their dynamics show a clear
\emph{separation of timescales}. 

For small times $t=\Theta(1)$, the student remains close to its random
initialization (cf.\ \cref{fig:vis}(a); we will establish these timescales in our
derivations of the next sections). At $t=\Theta(d/\eta)$ (blue in
\cref{fig:dynamics_closure_decomp}), the loss starts to decrease as $Q_1$ increases; $Q_1$ then converges to a fixed
value for the remainder of training. Subsequently, $Q_2$ evolves and converges to
its final value. This suggests that the early stages of learning may be described
by $Q_1$ and $Q_2$ alone; we will derive such a description in
\cref{sec:gaussian-closure}. The higher-order parameters $Q_p$
start moving after $Q_1$ and $Q_2$, and unlike the first- and second-order
modes, their dynamics are strongly coupled. Their concurrent growth accompanies the onset of specialization,
as measured by the increase in $Q_{W_\ell}^{01}$.
At this stage, student neurons become
strongly correlated with teacher neurons. 

We can gain further insight into the learning dynamics by considering the
\emph{distribution of teacher--student overlaps}. In \cref{fig:op-histogram}, we plot a histogram
of the entries of the teacher--student overlap taken from the simulations shown
in \cref{fig:dynamics_closure_decomp} at four representative time steps. Close
to initialization and early during training, we observe that the overlaps follow
a normal distribution. This suggests a Gaussian ansatz for the joint
distribution of a finite subset of $\Omega_{ij}$ that we formalize in
\cref{sec:gaussian-closure}. This joint distribution is characterized by the
empirical first two moments $Q_1$ and $Q_2$, and Wick's theorem expresses the
higher-order $\bOmega$ contractions in \cref{eq:Q_p} in terms of these two order
parameters. Specialization is marked by the breakdown of this normal
distribution: as the higher-order moments $Q_{p\geq 3}$ grow, some
teacher--student overlaps become large and the Gaussian closure in terms of
$Q_1$ and $Q_2$ fails (third column of \cref{fig:op-histogram}). The student hidden
neurons then break the rotational symmetry within the teacher subspace and align
with individual teacher neurons, as illustrated in \cref{fig:vis}(c). Once the
student--teacher matching stabilizes, the dynamics enter a final refinement
regime. The distribution of the matched diagonal overlaps converges to a delta
peak at one, while the off-diagonal overlaps retain an approximately Gaussian
background, as shown in the final column of \cref{fig:op-histogram} and illustrated
in \cref{fig:vis}(d).

\subsection{The Gaussian closure phase}%
\label{sec:gaussian-closure}

We now use these two observations to derive closed equations for $Q_1$ and,
under Gaussian closure, for $Q_2$.

\paragraph*{First-order flow: linear alignment.} At $t=\Theta(d/\eta)$, the dynamics are governed at leading order by the linear
Hermite component. For $p=1$, a scaling analysis shows that all higher-order
contractions in \cref{eq:Q_p} are subleading (\cref{Sec:flow_Q_1}), so the
first-order dynamics close without assuming that the entries of $\bOmega$ are
jointly Gaussian:
\begin{equation}
    \begin{aligned}
        Q_1^{01}(t+1)
        &= Q_1^{01}(t)
        + \frac{2\eta\mu_1^2}{d}
        (Q_1^{00}-Q_1^{01}), \\
        Q_1^{11}(t+1)
        &= Q_1^{11}(t)
        + \frac{4\eta\mu_1^2}{d}
        (Q_1^{01}-Q_1^{11}).
    \end{aligned}
    \label{eq:q1_dyn}
\end{equation}
Here, $\mu_1$ is the first Hermite coefficient of the activation function, and
$Q_1^{00}$ is fixed by the teacher. This two-dimensional linear system describes
the evolution of $Q_1$ from its initial values and agrees with simulations, as
shown in \cref{fig:relu_dyn_kd}.

\Cref{eq:q1_dyn} has a simple interpretation. The teacher--student
overlap $Q_1^{01}$ is driven by an effective attractive force toward
$Q_1^{00}$, which tends to align the linear component of the student network with the
teacher's. Similarly, the attraction of
$Q_1^{11}=\frac{1}{kd}\|\mathbf{1}^\intercal\bW\|^2$ toward $Q_1^{01}$ plays
the role of a regularization that keeps the norm of the student's linear
component from exceeding its overlap with the teacher's. The spherical normalization of the weights
only contributes a correction of order $O(\eta/d^2)$ at this stage and
therefore does not modify the leading-order evolution of $Q_1$. The first-order
flow connects our description to the equations of
\citet{saad1995line,saad1995exact} for soft committee machines. Specifically, our
\cref{eq:q1_dyn} follows from eq.~(14) of \citet{saad1995line} after translating
to our normalization, summing over hidden-neuron indices, and retaining the
leading linear Hermite contribution in the extensive-width limit. The main
difference is that Saad--Solla-type analyzes work at fixed hidden width, where
the full pairwise overlap dynamics are finite-dimensional and closed. In the
extensive-width limit, the overlap matrix itself becomes high-dimensional, and
we instead project it onto a moment hierarchy. Within this hierarchy, $Q_1$
aggregates the Saad--Solla overlaps over all pairs of neurons, and for
$\mu_2\neq0$, $Q_2$ adds a second closed stage. For odd activations such as erf or
tanh (also used by Saad and Solla), $\mu_2=0$, so this intermediate stage is
absent and the dynamics couple directly to higher-order sectors and
specialization, as shown for tanh in \cref{fig:dynamics_closure_decomp}. We now
derive this second-order flow.

\paragraph*{Second-order flow via Gaussian closure.}

\begin{figure}[t!]
  \centering
  \includegraphics[width=\linewidth]{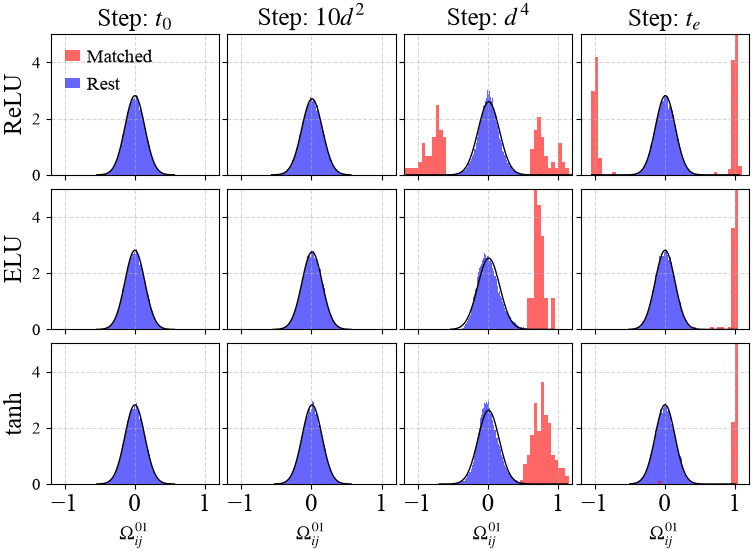}
  \caption{\label{fig:op-histogram} \textbf{Empirical distribution of the
    student--teacher overlap matrix entries at representative stages of
    training.} Histograms of the teacher--student overlaps
    $\Omega^{01}_{ij}$ at representative stages of training. Early on (first
    two columns), all entries lie within a Gaussian bulk (blue), whose mean and
    width are determined by $Q_1$ and $Q_2$. As training progresses, student
    neurons become strongly correlated with individual teacher neurons. For each
    student neuron, we highlight in red all overlaps whose magnitudes exceed the
    extremal value of $k$ random teacher--student overlaps, which scales like
    $\sqrt{2 \log k}$. Their appearance marks the emergence of specialization.
    The solid black curve is a Gaussian fit to the remaining overlaps.
    \emph{Parameters}: Width $k=200$ and input dimension $d=50$.}
\end{figure}

Once the linear mode has converged, the quadratic component drives the
dynamics, from $t=\Theta(d^2/\eta)$ until specialization. On this timescale
$Q_2$ obeys an autonomous flow. Starting from the master equations \cref{eq:Q_p}, we find that the
update of $Q_2$ involves loop and multipoint contractions of the
microscopic overlaps $\Omega_{ij}^{ab}$. At random initialization,
fixed finite collections of off-diagonal student--student overlaps, and any
teacher--student overlap, are asymptotically
jointly Gaussian. Motivated by this initial property and numerical
observations during early learning (\cref{fig:op-histogram}), we adopt a
\emph{Gaussian-closure} approximation for the overlap statistics before
specialization:
\begin{equation*}
    \big(\Omega_{i_1j_1}^{a_1b_1},\ldots,
    \Omega_{i_nj_n}^{a_nb_n}\big)
    \overset{\mathrm{closure}}{\sim}
    \mathsf{N}(\boldsymbol\mu,\boldsymbol\Sigma),
\end{equation*}
where $n$ is fixed and diagonal self-overlaps are treated separately.
The marginal means and variances are specified in \cref{Sec:GC}.  The correlation between distinct entries is set by the indices they
share: entries with no common neuron are uncorrelated at leading order,
while entries sharing one neuron, say $(a,i)$, have covariance
\begin{equation}
\operatorname{Cov}\!\big(\Omega^{ab}_{ij},\Omega^{ac}_{im}
    \,\big|\,\Omega^{bc}_{jm}\big)
    \approx\frac1d\,\Omega^{bc}_{jm}.
    \label{eq:local-cov}
\end{equation}
Wick's theorem then expresses the statistics of any fixed product of overlaps as a sum over
pairings determined by this index-sharing pattern. The overlaps generated by \cref{eq:local-cov} are subsequently
averaged using the same Gaussian-closure prescription. We further assume that this evaluation holds at leading order
for the normalized sums over extensively many entries as in \cref{eq:Q_p}.
The contractions in \cref{eq:Q_p} then depend only on $Q_1$ and $Q_2$. Keeping only the leading $O(\eta/d^2)$ terms yields the following closed
second-order flow,
\begin{align}
             Q_2^{01}(t+1) &= Q_2^{01}(t) +  \frac{4\eta\mu_2^2}{d^2}\Big[(Q_2^{00}-Q_2^{11})\nonumber\\
             &\quad+\frac{d}{k}Q_2^{01} (Q_{2}^{11} - Q_{2}^{01} )\Big],   \label{eq:q2_dyn}\\
             Q_2^{11}(t+1)  &=Q_2^{11}(t) + \frac{8\eta\mu_2^2}{d^2} \Big[2(Q_2^{01} - Q_2^{11}) \nonumber \\
             &\hspace{-1cm}+\frac{d}{k}(Q_2^{01}+Q_2^{00}-2Q_2^{11})+ \frac{d}{k}Q_2^{11} (Q_{2}^{11} - Q_{2}^{01}  ) \Big],\nonumber
        \end{align}
derived in \cref{Sec:Overlap_Flow,Sec:contractions}. Simulations again closely
match these predictions for $Q_2$ (\cref{fig:relu_dyn_kd}).

\begin{figure*}[t]
  \centering    
  \includegraphics[width=\linewidth]{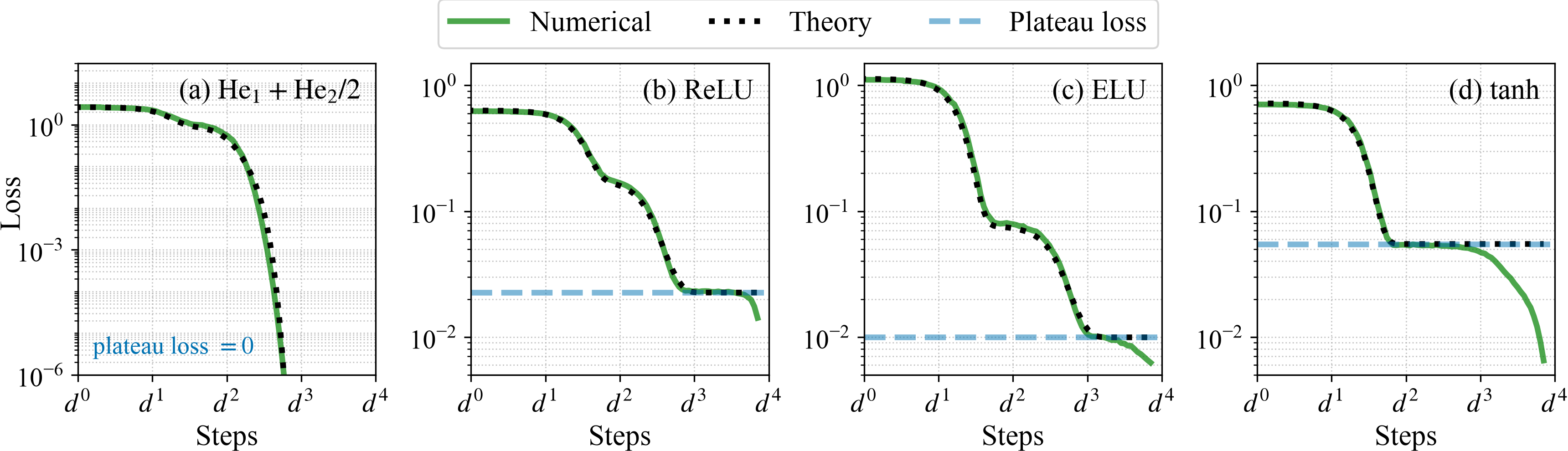}
    \caption{\label{fig:dynamics_closure} \textbf{Generalization loss dynamics
    under Gaussian closure.} We show the test loss \cref{eq:loss_decomp}
    obtained by simulating neural networks together with the prediction obtained
    under Gaussian closure, \cref{eq:loss_th}. Panel (a) uses the activation
    $\sigma(x)=\mathrm{He}_1(x)+\mathrm{He}_2(x)/2$, for which the predicted
    plateau loss is zero. In this case, Gaussian closure describes the entire
    training dynamics; panels (b--d) show ReLU, ELU, and tanh activation
    functions. The blue dashed line shows the theoretical pre-specialization
    loss plateau after removing the contributions from the first two Hermite
    sectors.  \emph{Parameters:}~$k=400,d=100$.}
\end{figure*}

Although more intricate than \cref{eq:q1_dyn}, \cref{eq:q2_dyn} similarly drives the teacher--student and
student--student overlaps to match the teacher--teacher overlap. As before, the
student--student term $Q_2^{11}(t)\propto \|\bW^\intercal \bW\|_{\rm F}^2$ is a
regularization that penalizes alignment between different student
neurons and therefore reduces feature redundancy. Unlike for $Q_1$, here the
spherical normalization of the weight update contributes at leading order
through the final terms in both equations. 

Combining the closed $Q_1$ and $Q_2$ flows yields the generalization loss
throughout the pre-specialization regime. We use the fact that terms
beyond the second-order one in the expansion of the teacher network are not yet learned and therefore play the role of
effective Gaussian noise, as in the static analysis of \citet{barbier2026statistical}. This resembles a form of Gaussian
equivalence, where a random function (here, the response of the teacher network)
is approximated by the first two terms of its Hermite expansion, while the
remaining terms are subsumed in an effective loss term~\citep{goldt2020modeling, gerace2020generalisation, goldt2022gaussian, mei2022generalization,hu2022universality}. Substituting $Q^{00}_{p\ge 3}=Q^{11}_{p\ge 3}=1$ and
$Q^{01}_{p\ge 3}=0$ into the loss decomposition in \cref{eq:loss_decomp} and using
the identity
$\mathbb{E}[\sigma^2]\coloneq \mathbb{E}_{z\sim\mathsf{N}(0,1)}[\sigma^2(z)]=\sum_{\ell\ge
0} \mu_\ell^2/\ell !$, we obtain
\begin{equation}
     \mathcal{L} =  \sum_{\ell=1}^{2}\frac{\mu_\ell^2}{\ell!}(Q_\ell^{00}-2Q_\ell^{01}+Q_\ell^{11}) + 2\Big(\mathbb{E}[\sigma^2] - \sum_{\ell=0}^2 \frac{\mu_\ell^2}{\ell!}\Big) \label{eq:loss_th}.
\end{equation}

The closed flow for $Q_1$ and $Q_2$ describes the learning dynamics of neural
networks with high accuracy, as shown in \cref{fig:dynamics_closure}. For a
quadratic activation function, this description remains accurate throughout
learning, capturing the entire loss trajectory down to zero. For the other
activation functions shown, it tracks the loss up to the plateau reached at the
end of the Gaussian-closure stage. By this point, the student has learned the
quadratic approximation of the teacher,
\begin{equation*}
    \sqrt{k}\mu_0
    + \frac{\mu_1}{\sqrt{kd}}
      \sum_{i=1}^{k}\mathbf w_i^0\cdot\bx
    + \frac{\mu_2}{2\sqrt{k}}
      \sum_{i=1}^{k}
      \Big[
      \Big(\frac{\mathbf w_i^0\cdot\bx}{\sqrt d}\Big)^2
      -1
      \Big],
\end{equation*}
which is precisely what a kernel or random-feature machine learns at quadratic
sample complexity~\citep{mei2022generalization, misiakiewicz2023six}. For activations with nonzero Hermite components beyond degree two, further learning proceeds through specialization, which breaks Gaussian closure and which we analyze next.

\subsection{The specialization transition}
\label{sec:specialization}

We now analyze the dynamics of specialization, that is the
emergence of macroscopic correlations between individual student and teacher
neurons that appear as outliers separating
from the bulk of overlaps in \cref{fig:op-histogram}. 

An exact,
low-dimensional description of this regime analogous to Gaussian closure is
currently out of reach. Here, we therefore focus on estimating the
onset of specialization: we estimate the time required for a representative
teacher--student overlap $\omega \coloneq \Omega^{01}_{ij}$ to escape the bulk (where $\omega 
\asymp d^{-\nicefrac{1}{2}}$) and become macroscopic ($\omega =\Theta(1)$).
This approach parallels the analysis of weak recovery of teacher
features in single- and multi-index models; see, for example,
\citet{benarous2021online}. The key difference in our case is that the network
has already learned the linear and quadratic components of the teacher before
specializing; we
will show that this impacts the effective single-neuron dynamics in several
ways. We delegate detailed derivations to \cref{app:sec:specialization}.
 
\paragraph*{The timescale of specialization.}
Our starting point is an effective equation of motion for $\omega$ on the
\emph{Gaussian plateau}, i.e.\ after $Q_1$ and $Q_2$ have converged to their
asymptotic value. Having learned the linear and quadratic parts of the teacher
means that the first two terms of the loss
decomposition~\cref{eq:loss_decomp} vanish, i.e.\ that
\begin{equation}
    \label{eq:gc_order_param_constraint}
    Q_p^{00}-2Q_p^{01}+Q_p^{11}=0 \quad \text{for}\quad p=1,2.
\end{equation} 
In the master
equations~\cref{eq:Q_p}, these constraints imply that the drift terms
proportional to $\mu_1^2$, $\mu_2^2$, $\mu_1\mu_3$ and $\mu_2\mu_4$ vanish. On
the Gaussian plateau, the growth of $\omega$ is thus driven
by the $r$th Hermite mode; $r$ is thus the specialization exponent
\begin{equation}
    \label{eq:spec_exp_def}
    r=r(\sigma):=\min\{s\geq3:\mu_s(\sigma)\neq0\},
\end{equation}
that is the degree of the term in the Hermite expansion of $\sigma$ that will
drive specialization. We are therefore led to consider the effective activation
function
\begin{equation}
  \sigma(z) = \mu_0+\mu_1 z+\frac{\mu_2}{2}\,(z^2-1)+\varepsilon\,\psi_r(z),
\end{equation}
where $\psi_r(z)\coloneq\mathrm{He}_r(z) / \sqrt{r!}$ is normalized w.r.t.\ the standard Gaussian measure, and $\varepsilon\coloneq
\mu_r/\sqrt{r!}$.
 
Expanding a single SGD step including normalization to second order in $\eta$ and averaging over the fresh sample
under Gaussian closure yields an effective equation of motion in terms of the orders:
\begin{equation}
  \label{eq:specialization_eom}
  \mathbb{E}[\omega(t+1)-\omega(t)]\asymp 
\frac{2r\eta\varepsilon^2}{kd}\omega^{r-1}
-\frac{2\eta^2\varepsilon^2\mathcal{L}_ra_\sigma}{kd}\omega,
\end{equation}
with $a_\sigma\coloneq \mathbb E_{z\sim
\mathsf{N}(0,1)}[\sigma'(z)^2]=\mu_1^2+\mu_2^2+r \varepsilon_r^2$ and
$\mathcal{L}_r\coloneq Q_r^{00}-2Q_r^{01}+Q_r^{11}=\Theta(1)$, so that $\mu_r^2\mathcal{L}_r/r!$ is
the leading order of the loss on the plateau. The first term in \cref{eq:specialization_eom} is a drift and describes alignment of the student neuron with the teacher neuron and is driven by the population gradient. The second term, in $\eta^2$, is a linear shrinkage induced by the normalization. Its magnitude depends on the second moment of the sample gradient, and therefore also reflects gradient fluctuations. The coupling to other neurons plays a role too in the dynamics, but does not change the order of $\omega$ in~\cref{eq:specialization_eom}, see \cref{app:E1,app:E2}.

To estimate the onset of specialization, we track the largest overlap
of the student neuron with some teacher neuron. By virtue of being the largest,
its initial value~$\omega_0$ follows the extreme-value statistics of the
Gaussian distribution of overlaps on the plateau, which has mean~$Q_1^{01}/k$ and second moment $Q_2^{01}/k$, so
\begin{equation*}
  \omega_0\asymp\sqrt{\frac{2Q_2^{01}\log k}{k}}
  =\Theta\Big(\sqrt{\frac{\log k}{k}}\Big).
\end{equation*}
Neglecting the shrinkage for now, 
the leading overlap follows the gradient flow
$\dot\omega=\Theta(\eta r \varepsilon^2\omega^{r-1}/kd)$ (using the dot notation for a time derivative). Integrating from~$\omega_0$ to
a fixed threshold $\omega_{\rm spec}=\Theta(1)$ gives a gradient-flow
specialization timescale
\begin{equation}
  \label{eq:tau_spec_gf}
  \tau_{\rm GF}
  =\Theta\Big(\frac{1}{\eta \varepsilon^2}\frac{k^{1+r/2}}{(\log k)^{(r-2)/2}}\Big).
\end{equation}
Once a finite fraction of the leading overlaps has reached $\omega_{\rm
spec}$, $Q_{W_\ell}^{01}$ increases by a finite amount, so
\cref{eq:tau_spec_gf} is the specialization time at fixed learning
rate.
 
The result \cref{eq:tau_spec_gf} is for gradient flow and ignores the effect of
the SGD noise. The noise enters the shrinkage term, which in turn bounds the learning rate. At the start of
the plateau, when the noise term is most relevant, balancing the two terms of
\cref{eq:specialization_eom} for $\omega = \omega_0$ gives the largest admissible learning
rate $\eta_c$; 
at a fixed fraction of $\eta_c$,
\cref{eq:tau_spec_gf} becomes
\begin{equation}
  \label{eq:tau_spec_sgd}
  \tau_{\rm spec}=\Theta\Big(
\frac{a_\sigma \mathcal{L}_r}{\varepsilon^2 r}
\frac{k^r}{(\log k)^{r-2}}
\Big).
\end{equation}
Specialization in an extensive-width network thus requires a number of samples
that grows as the $r$th power of the width (equivalently, of the input
dimension, since $k/d$ is fixed). For a single student neuron trained on a
single teacher neuron with information
exponent $r > 2$, online SGD needs $\Theta(d^{r-1})$ samples up to
logarithmic factors~\citep{benarous2021online}. \Cref{eq:tau_spec_sgd}
contains an additional factor of~$k$. This factor of $k$ is an extensive-width
effect: a student neuron has to recover a single teacher neuron from an
extensive number of unresolved teacher neurons using only the teacher output;
likewise, each student neuron only makes a small contribution to the student
output. The logarithmic correction is a direct consequence of the
extreme-value statistics at initialization induced by the extensive width of the student, and enters once through the time
integral and once through $\eta_c$. 

\begin{figure}[t!]
  \centering    
  \includegraphics[width=\linewidth]{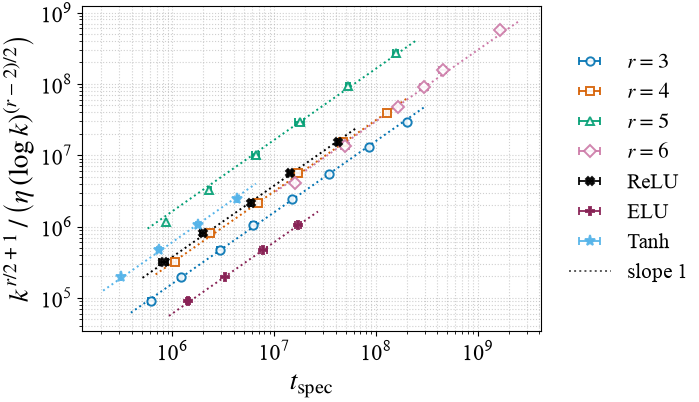}    
  \caption{\label{fig:timescale_exp-vs-theo}\textbf{Timescale of specialization.} We plot the time of specialization measured
  experimentally against the theoretical prediction,
  \cref{eq:tau_spec_gf}. Symbols indicate the SGD step where networks
  with varying $k=d$ specialize, defined as the first time at which
  $Q_{W_\ell}^{01}$ deviates from its Gaussian-closure value by more than $0.1$,
  with $\ell=1$ for odd $r$ and $\ell=2$ for even $r$. The detailed data
  collection procedure is illustrated in \cref{fig:specialization_trajs}. We fix the learning
  rate for all networks at $\eta=0.1$; the activation functions for $r=3,4,5,6$
  are defined as $\sigma_r(z) =
  \mathrm{He}_1(z)+\frac{1}{2}\mathrm{He}_2(z)+\frac{\mu_r}{r!}\mathrm{He}_r(z)$ with
  $\mu_r=0.25,0.4,1.2,0.72$, respectively.}
\end{figure}

To verify our prediction experimentally, we measured the onset of specialization
during online SGD in neural networks with standard activation functions like
ReLU, ELU, tanh, and others. In \cref{fig:timescale_exp-vs-theo}, we see that
during training at fixed learning rate $\eta = 0.1$, 
the experimentally measured onset of specialization follows precisely the
\emph{gradient flow} prediction of \cref{eq:tau_spec_gf} for different
specialization exponents. The largest networks tested reach $k=d=560$. In
other words, online SGD specializes as fast as the noise-free gradient flow
in a practical setting without having to scale down the learning rate with the dimension. We now explain that a noise-reduction effect that is specific to the extensive-rank setting is responsible for the apparent speed-up; we call this effect \emph{quadratic shielding}.

\paragraph*{Quadratic shielding reduces SGD noise.} Our estimate of
$\tau_\mathrm{spec}$ in \cref{eq:tau_spec_sgd} accounts for the learned quadratic
part only by assuming that the first two Hermite modes do not contribute to the
loss, without considering its effect on the weights. In fact, rewriting the
constraints from \cref{eq:gc_order_param_constraint} in terms of the weights, we
find that learning the quadratic part of the teacher function requires the
student weights to
match the first and second-order statistics of the teacher weights:
\begin{equation}
  \label{eq:low_order_matching}
  \sum_{i=1}^k\mathbf w_i=\sum_{i=1}^k\mathbf w_i^0
  \quad \text{and} \quad
  \bW^\intercal\bW=\bW^{0\intercal}\bW^{0}.
\end{equation}
The second condition fixes $\bW$ only up to an orthogonal transformation in
neuron space, $\bW=\mathbf{O}\bW^0$, which is why the individual overlaps
$\Omega^{01}_{ij}$ remain 
at the scale $k^{-1/2}$ although the
quadratic part of the teacher has been learned. This effective rotational symmetry of the weights is the source of the so-called ``universal phase'' in the static analysis of \citet{barbier2026statistical} which, as in the present paper, precedes specialization as data size grows. As we now argue, these constraints 
reduce the effective noise of SGD compared to its strength before the
Gaussian-plateau.

On the Gaussian plateau, simply discarding the learned low-degree components
leaves the effective activation $\sigma_{\rm eff}=\varepsilon\,\psi_r$. For this reduced activation, the
population gradient would still scale like $\varepsilon^2$, while its variance would decrease
to only $\varepsilon^4$, suggesting weaker stochastic fluctuations and a larger admissible learning rate than the
estimate above. However, this simplification ignores that even
after learning the linear and quadratic part of the teacher, the first two Hermite modes still contribute to the noise, yielding an instantaneous noise
whose variance is proportional to $\varepsilon^2$. As far as specialization is concerned, the
low-order parts of the teacher output hence act like noise, obfuscating the
signal coming from~$\psi_r$.

Quadratic shielding suppresses the accumulated effect of these fluctuations on specialization. The dynamics of $Q_1$ and $Q_2$, which occur on timescales~$d/\eta$ and~$d^2/\eta$ that are
much shorter than $\tau_{\rm spec}$ (cf.\ \cref{sec:observations}), effectively project the weights onto the manifold $\mathcal{M}$ defined by the
constraints \cref{eq:low_order_matching}. Indeed, we show in
\cref{app:sec:quadratic-shielding} that confining the weights explicitly to $\mathcal{M}$ restores the
noise scale of the pure degree-$r$ problem: the drift remains proportional
to $\eta\varepsilon^2$, while the noise decreases to be proportional to $\varepsilon^4$. Balancing drift and noise again then gives gives the largest admissible learning rate $\eta_{c,\mathcal M}$ 
and a specialization timescale of
\begin{equation}
  \label{eq:tau_spec_shielded}
  \tau_{\rm spec,\mathcal{M}}=\Theta\Big(
\frac{\mathcal{L}_r}{r}
\frac{k^r}{(\log k)^{r-2}}
\Big).
\end{equation}
Therefore the admissible learning rate increases, and the specialization time
decreases, by a factor $a_\sigma /\varepsilon^2$, without changing the
dependence on the dimension. For the ReLU activation, $a_\sigma /\varepsilon^2
\approx 75$. At fixed $\eta$, this suppression instead extends the regime
in which gradient flow accurately describes specialization, providing a possible
explanation for the agreement in \cref{fig:timescale_exp-vs-theo}. 

We emphasize however that at the level of our derivation, the confinement of the
weights to $\mathcal M$ is an assumption: SGD does not enforce
\cref{eq:low_order_matching} sample by sample. We did however test this mechanism by
measuring the mean-square displacement (MSD) of teacher--student overlaps on the
Gaussian plateau, see \cref{fig:quadratic-shielding}. The slope of the MSD
crosses from an $\varepsilon^2$ dependence at short lags to an $\varepsilon^4$
dependence for lags that are larger than the timescale of the $Q_2$ dynamics
$\tau_{Q_2}$. So although the instantaneous noise remains quadratic in
$\varepsilon$ due to the ``cross-talk'' from the linear and quadratic part of
the teacher at short timescales, its contribution to the overlaps is suppressed by the quadratic
shielding from the $Q_2$ dynamics at specialization timescale. This observation supports the
effective-confinement approximation.

\paragraph*{Three scenarios for specialization.}%

The different patterns of specialization that we see in the third column of
\cref{fig:op-histogram} depend on the activation function, and more precisely on
on which of its Hermite coefficients are nonzero. We distinguish three
scenarios.

\begin{figure*}[t!!]
    \centering    \includegraphics[width=\linewidth]{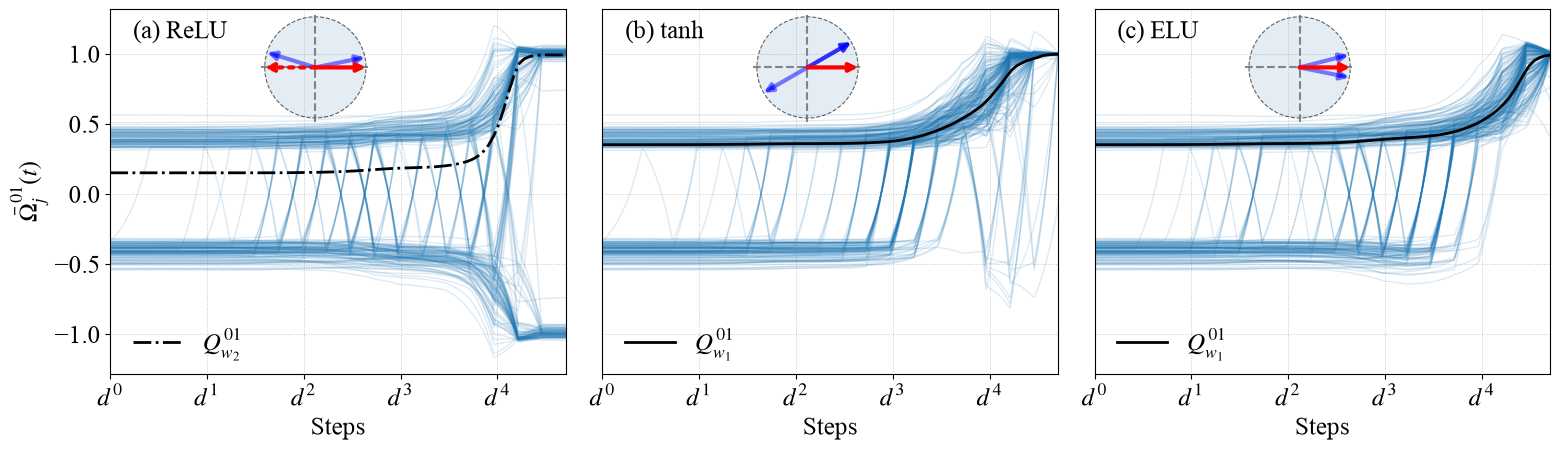}
    \caption{\label{fig:specialization-scenarios} \textbf{Three scenarios of
    specialization.} We
    illustrate the dynamics of specialization for three representative
    activation functions.  For each student neuron, we show the largest
    alignment it has with a teacher neuron (thin blue curves): at each update
    $t$ and for each student neuron $j$, we set $\bar \Omega_j^{01}(t)\coloneq \Omega_{i'j}^{01}(t), i'=\arg
    \max_{i\in[k]}|\Omega_{ij}^{01}(t)|$. The black curves show the corresponding
    macroscopic order parameter defined by \cref{eq:Q_w}. Insets schematically illustrate the three specialization patterns
    discussed in the text: (a) sign-degenerate matching for activations with only
    even nonlinear Hermite components, such as ReLU $(r=4)$;
    (b) student--student anti-alignment
    for activations with only odd nonlinear Hermite components, such as tanh $(r=3)$; (c) many-to-one matching without a sign symmetry for mixed-parity
    activations, such as ELU $(r=3)$.  \emph{Parameters: $d=50$, $k=200$, and $\eta=0.1$.}}
\end{figure*}

For ReLU-like activations (ReLU, GELU, Swish, PReLU, leaky ReLU, Softplus), the
Hermite expansion contains only even terms beyond the linear one. Hence a student neuron can specialize either by aligning or by
anti-aligning with a teacher neuron: both contribute equally to the leading
nonlinear term of the dynamics and the loss \cref{eq:loss_decomp}. \Cref{fig:specialization-scenarios}(a) shows
this: the largest overlaps (light blue) are bimodal both before
specialization, when their signs are set by the initialization, and after it.
This behavior is consistent with the overlap distributions in
\cref{fig:op-histogram}, where the specialized ReLU features split into strongly
positive and strongly negative overlap modes. The even-order overlaps, such as $Q_{W_2}^{01}$, can therefore
grow and amplify the even higher-order parameters, whereas the odd moments,
like $Q_{W_1}^{01}$ and $Q_{3}^{01}$, 
remain suppressed. With fixed second-layer weights, however, the student can learn
the teacher perfectly only if its neurons align, rather than anti-align, with the
teacher neurons. ReLU-like students therefore remain stuck in suboptimal
configurations, as reported for ReLU networks by \citet{safran2018spurious}, who
also showed that mild overparameterization can overcome this bottleneck.

Activations like tanh, erf, and Softsign instead have only odd Hermite
components. During the transition, student neurons
therefore correctly align with the teacher neurons, see the
specialization trajectories in \cref{fig:specialization-scenarios}(b) and the
histogram in \cref{fig:op-histogram}. In tanh-like networks, some student
neurons also anti-align with each other during the transition. Since tanh is
odd, such a pair cancels in the output and effectively drops out, which can
lower the test loss. This shows up as a transient growth of the even-order
self-overlaps $Q^{11}_{p}$ (fourth row of \cref{fig:dynamics_closure_decomp}),
although the instability itself is driven by the odd mode $Q_3$.
We sketch this behavior in the inset of \cref{fig:specialization-scenarios}(b).
However, this anti-alignment is transient, and ultimately all student neurons
correctly align with their corresponding teacher neuron.

Finally, activations such as ELU have both odd and even nonlinear Hermite
components; the cubic mode is then typically the first higher-order mode to
become unstable. Because both parities are active, students align correctly with their
teacher neurons, and do not anti-align with each other, as shown in
\cref{fig:specialization-scenarios}(c). 

Specialization need not produce a one-to-one matching at once:
early on, several student neurons can align with the same teacher one.
Similar many-to-one specialization patterns have been observed in
over-realizable soft committee machines and overparameterized teacher--student
networks~\cite{goldt2019dynamics,tian2020student, richert2022soft}. Such
configurations indicate that specialization has begun, while the final permutation-like student--teacher assignment remains unresolved.

\subsection{The refinement phase}%
\label{sec:refinement}

In the \emph{refinement phase} in the long-time regime after specialization,  the student has converged onto a
specific matching between student and teacher neurons. Without loss of
generality, we relabel the neurons so that student $i$ is matched to teacher $i$,
with matched overlap $\Omega_{ii}^{01}$. The matched diagonal overlaps are $\Theta(1)$, while the
unmatched off-diagonal overlaps remain at the Gaussian background scale,
$\Omega_{ij}^{01}=O(k^{-1/2})$ for $i\ne j$. Therefore, the higher-order
moments are dominated by the diagonal: $Q_p^{01}\approx
k^{-1}\sum_{i=1}^k(\Omega_{ii}^{01})^p$ for $p\ge 3$, similarly to the
equilibrium analysis of \citet{barbier2026statistical}. The late-time overlap
histogram in \cref{fig:op-histogram} bears out this structure.

The diagonal overlaps are determined by the stationary points of the local
diagonal flow derived in \cref{sec:self-consis}. At leading order they satisfy, for all $i\in[k]$,
\begin{equation}
    \begin{aligned}
           &\Omega_{ii}^{01}(t+1) =\Omega_{ii}^{01}(t)\\
           &\qquad+\frac{2\eta}{kd}\sum_{\ell\ge2}
    \frac{\mu_{\ell+1}^2}{\ell!}
    \Big[
        (\Omega_{ii}^{01})^\ell
        -
        \Omega_{ii}^{01}
        \frac{1}{k}
        \sum_{s=1}^k
        (\Omega_{ss}^{01})^{\ell+1}
    \Big].
    \label{eq:diag_Q1_main}
    \end{aligned}
\end{equation}
The $\ell=0,1$ contributions are absent from \cref{eq:diag_Q1_main}, consistent with the picture developed above: $Q_1,Q_2$ control the collective evolution of the student--teacher overlap matrix, but they do not resolve the individual student--teacher assignment. After $Q_1$ and $Q_2$ learned, these sectors do not contribute to the leading order stationary diagonal equation.

The nonzero stationary points of \cref{eq:diag_Q1_main} determine the stationary
distribution of student--teacher overlaps. We find them by imposing
stationarity mode by mode: for every $\ell\ge2$ with $\mu_{\ell+1}\ne0$, the nonzero
sector-wise condition is
$(\Omega_{ii}^{01})^{\ell-1}=k^{-1}\sum_{s=1}^k(\Omega_{ss}^{01})^{\ell+1}$ for
$i=1,\dots,k$. For activations with an active odd nonlinear Hermite component,
such as tanh or ELU, this condition gives the specialized fixed
point $\Omega_{ii}^{01}=1$ for all $i$.
Consequently $Q_{W_1}^{01}=1$ and $Q_p^{01}=1$ at leading order for $p>2$. The
corresponding stationary diagonal law is
$k^{-1}\sum_i\delta_{\Omega_{ii}^{01}}\rightarrow\delta_1$, and the full overlap
distribution is this diagonal atom plus the off-diagonal Gaussian background. 

For rectified activations such as ReLU, all nonlinear odd Hermite components are
absent. The reduced dynamics in \cref{eq:diag_Q1_main} therefore admits
specialized stationary configurations with $\Omega_{ii}^{01}\in\{-1,+1\}$
for all $i$, allowing arbitrary assignments of the two signs. The stationary diagonal law
may therefore contain both signs,
$k^{-1}\sum_i\delta_{\Omega_{ii}^{01}}\rightarrow\rho_+\delta_1+\rho_-\delta_{-1}$
with $\rho_++\rho_-=1$. The weights $\rho_+$ and $\rho_-$ are not fixed by
\cref{eq:diag_Q1_main}; they depend on the earlier specialization dynamics and
on the initialization. Equivalently, the full overlap distribution has a
two-peak diagonal component plus an off-diagonal Gaussian background. This
refinement picture is consistent with the specialization patterns discussed
above and with the overlap distributions in the last column of
\cref{fig:op-histogram}. In the rectified case, $Q_{W_2}^{01}=1$ and the even
moments satisfy $Q_p^{01}=1$ at leading order, as shown in the last row of
\cref{fig:dynamics_closure_decomp}. 

\section{Discussion}%
\label{sec:discussion}

We have revisited the Saad--Solla equations for the learning dynamics of
two-layer neural networks in the extensive-width limit, where the hidden width
and input dimension grow proportionally. This approach first changes variables
from the student weights to their overlaps with the teacher weights. For the narrow networks considered by Saad and Solla, this
yields an $\Theta(1)$ number of order parameters whose dynamics can be described
exactly, but at extensive width, we are left with an extensive number of
overlaps with no obvious way to further reduce the dimensionality of the
description.

We therefore took a ``mesoscopic'' perspective on these order parameters by
seeking to describe the learning dynamics via the \emph{distribution} of
overlaps. In the early stages of learning, we found that a Gaussian assumption
on the joint distribution of overlaps yields a closed flow for low-order overlap
statistics which accurately describes the learning up to teacher's quadratic
component.  Specialization marks the breakdown of this closure, when an
extensive number of student neurons strongly correlate with their teacher
neurons and the overlap distribution splits into a bulk and an extensive number
of outliers. We estimated the timescale of specialization and found a   quadratic shielding mechanism, whereby learning the linear and
quadratic components of the target function filters gradient fluctuations and
accelerates learning. 

The concept of Gaussian closure may seem reminiscent of the Gaussian equivalence
principle for neural networks and random feature
models~\citep{goldt2020modeling, goldt2022gaussian, gerace2020generalisation,
loureiro2021learning, mei2022generalization, hu2022universality,
montanari2022universality}. The two notions are, however, different. For narrow
two-layer networks trained on structured inputs, Gaussian equivalence
establishes that the joint distribution of local pre-activations is jointly
Gaussian~\citep{goldt2020modeling, goldt2022gaussian, hu2022universality,
loureiro2021learning, montanari2022universality}. Similar
Gaussian-equivalence descriptions also simplify the characterization of
generalization error of random features~\citep{gerace2020generalisation,
mei2022generalization, hu2022universality}. Here, we instead assume that a
finite subset of teacher--student overlaps behave as jointly Gaussian random
variables along the early portion of the SGD trajectory. Its breakdown signals
the onset of specialization.

Interestingly, the leading-order scaling of the specialization time can be
obtained from an effective single-neuron description. However, we do not recover
the sample complexity of single-index models: our estimate retains an additional
factor of $k$ which reflects the difficulty of recovering a specific teacher neuron
from an extensive-rank network. Another difference to single-index
models is that the network already learns non-trivial structure from the teacher
before specialization, even though individual teacher features remain
unresolved. This collective learning accelerates specialization through
quadratic shielding, which reduces the cumulative effect of large instantaneous
gradient fluctuations. The extensive width also introduces an
extreme-value effect: each student neuron can specialize towards one of many
teacher neurons, so its largest initial overlap is enhanced by a factor of
$\sqrt{\log k}$ relative to a typical overlap.

A key assumption of our work is that each SGD step uses a fresh sample, so
predictions like the specialization timescale crucially depend on it. While it
was initially conjectured that similar predictions in the online setting for
learning timescale based on information or leap exponents would extend to SGD
with re-use, \citet{dandi2024benefits} and \citet{arnaboldi2024repetita} showed
that reusing samples allows a two-layer network to learn directions with
information exponent larger than two
Studying the dynamics of specialization for SGD with re-use is therefore an
important future direction to establish whether there is a simple invariant that
predicts the specialization timescale for this algorithm, akin to the
\emph{generative exponent} of \citet{damian2024computational}. 

A natural question is whether we can complete the ``mesoscopic'' picture by
finding a simple description of the overlap distribution beyond specialization
that would reduce the number of necessary order parameters, akin to Gaussian
closure in the early stages of learning. Only $O(k)$
of the $O(k^2)$ overlaps detach from the bulk. We found that a
naive ansatz consisting of a Gaussian bulk and a single evolving atomic spike
for the $O(k)$ outliers already captures the dynamics of a network starting from
an ``informed'' initialization, where each student neuron has a small but finite
alignment with a specific teacher neuron at initialization, see
\cref{fig:informed-init}. The resulting dynamics close on five macroscopic order
parameters and reproduce the observed training trajectory, but they
miss the specialization transition by construction. Whether specialization can be captured by a
small number of evolving distributional parameters remains an appealing open
question.

\section{Concluding perspectives}

An important next step is to extend the mesoscopic description of learning
dynamics to mismatched teacher--student architectures, and to deep networks, for example along the lines of
\citet{barbier2026statistical}. Equally important is understanding how the
structure of the data shapes learning in extensive-rank networks that are able
to capture more of that structure than narrow networks. Hidden or neural
manifold models~\citep{chung2018classification, cohen2020separability,
goldt2020modeling}, hierarchical models~\citep{cagnetta2024deep} or explicitly
non-Gaussian models of data~\citep{ingrosso2022data, bardone2024sliding,
lufkin2024nonlinear} would
allow us to test which aspects of collective learning and neuron specialization
persist beyond isotropic Gaussian inputs.

A broader challenge is to identify signatures of these stages when the teacher
is unknown, from the learned representations and network responses alone. Finally, a fundamental
question that emerges from our dichotomy between Gaussian closure and
specialization is when the acquisition of useful representations in systems with
many trainable degrees of freedom can be understood through a small number of
order parameters, and when the emerging dynamics are irreducibly complex.

\section*{Acknowledgments}

The authors gratefully acknowledge funding by the European Union (ERC BEYOND2,
project 101166056 for S.Z. and S.G., and ERC CHORAL, project 101039794 for
J.B.). Views and opinions expressed are however those of the authors only and do
not necessarily reflect those of the European Union or the European Research
Council. Neither the European Union nor the granting authority can be held
responsible for them. S.Z.\ gratefully acknowledges the Abdus Salam International Centre for Theoretical Physics in Trieste for its hospitality during a visit when this
work was initiated, as well as Rudy Skerk for valuable discussions and help
with the code.

\bibliographystyle{apsrev4-2}
\bibliography{ref.bib}

\clearpage

\appendix
\onecolumngrid


\section{Gaussian closure}
\label{Sec:GC}

By \emph{Gaussian closure} we mean the following
approximation, used before specialization. Finite collections of off-diagonal
overlap entries $\bOmega$ are treated as jointly Gaussian, with covariances
obtained from Wick contractions of the weights $\bW^0$ and $\bW$, whose
coordinate moments are taken to be Gaussian at leading order. Diagonal
self-overlaps are set to one. After the index sums, every retained contraction
in the order-parameter dynamics is then a function of $Q_1$ and $Q_2$.

More precisely, in this regime, for any fixed finite collection of entries taken from
\begin{equation*}
    \bOmega^{00}-\operatorname{diag}(\bOmega^{00}),
    \qquad
    \bOmega^{01},
    \qquad
    \bOmega^{11}-\operatorname{diag}(\bOmega^{11}),
\end{equation*}
namely $\{
        \Omega_{i_sj_s}^{a_sb_s}
    \}_{s=1}^n$ where $n$ is fixed independently of $k,d$, and where $i_s\neq j_s$ for
the self-overlap blocks $a_s=b_s$, the following vector is assumed to be approximately jointly Gaussian under
the closure:
\begin{equation*}
    \big(
        \Omega_{i_1j_1}^{a_1b_1},
        \ldots,
        \Omega_{i_nj_n}^{a_nb_n}
    \big)
    \overset{\mathrm{closure}}{\sim}
    \mathsf{N}(\boldsymbol\mu,\boldsymbol\Sigma).
\end{equation*}
The empirical marginal mean and variance are fixed by
the first two empirical overlap moments. Correlations
between distinct entries additionally depend on their index-sharing
structure, as described below. For the teacher--student block
$\bOmega^{01}$, the entrywise empirical mean and variance are
\begin{equation*}
    \mu^{01}
    =
    \frac{Q_1^{01}}{k},
    \qquad
    (\sigma^{01})^2
    =
    \frac{Q_2^{01}}{k}
    -
    \Big(
        \frac{Q_1^{01}}{k}
    \Big)^2.
\end{equation*}
For the off-diagonal part of the $aa$-block, $a\in\{0,1\}$, the corresponding mean and variance are
\begin{equation*}
    \mu_{\mathrm{off}}^{aa}
    =
    \frac{Q_1^{aa}-Q_{W_1}^{aa}}{k-1},
    \qquad
    (\sigma_{\mathrm{off}}^{aa})^2
    =
    \frac{Q_2^{aa}-Q_{W_2}^{aa}}{k-1}
    -
    \Big(
        \frac{Q_1^{aa}-Q_{W_1}^{aa}}{k-1}
    \Big)^2.
\end{equation*}
Here $Q_{W_\ell}^{aa}\coloneq \frac1k\sum_{i=1}^k(\Omega^{aa}_{ii})^\ell$, $\ell=1,2$,
denote the first two diagonal self-overlap moments; for the self-blocks this
replaces the maximum over permutations in \cref{eq:Q_w}. We use $Q_{W_1}^{aa}=1$ and
$Q_{W_2}^{aa}\approx1$. 


The covariance $\boldsymbol\Sigma$ is obtained from Wick contractions; for
identical entries it reduces to the variance above. If two
entries share no network-labeled neuron, we take their
leading covariance to be zero:
\begin{equation*}
    \operatorname{Cov}
    \big(
        \Omega_{ij}^{ab},
        \Omega_{ml}^{ab}
    \big)
    \approx 0,
    \qquad
    \{(a,i),(b,j)\}\cap\{(a,m),(b,l)\}=\varnothing.
\end{equation*}
If two entries share one index, the local covariance is given by the
corresponding Wick contraction. For example (see
\cref{subsec:Three-Point}), contracting the shared weight
vector with unit coordinate covariance gives the local Wick estimate
\begin{equation*}
    \operatorname{Cov}_{\mathbf{w}^a_i}
    \big(
        \Omega_{ij}^{ab},
        \Omega_{im}^{ab}
        \mid
        \mathbf{w}_j^b,\mathbf{w}_m^b
    \big)
    \approx
    \frac1d\Omega_{jm}^{bb},
\end{equation*}
where $\mathbf w_i^a\in\mathbb R^d$ denotes the $i$th row of $\bW^a$
and is distinct from both conditioned neurons. This
local covariance may still depend on the realized background overlap
$\Omega_{jm}^{bb}$. Averaging this background factor
over all $j,m$ gives
\begin{equation*}
    \frac{1}{k^2}\sum_{j,m}\frac{\Omega_{jm}^{bb}}{d}
    =\frac{Q_1^{bb}}{dk}.
\end{equation*}

More generally, for any fixed contraction pattern appearing in the
macroscopic dynamics,
\begin{equation*}
    C_{k,d}
    \sum_{i_1,\ldots,i_n}\sum_{j_1,\ldots,j_n}
    \prod_{e=1}^n
    \Omega_{i_ej_e}^{a_eb_e},
\end{equation*}
with the index identifications specified by that pattern,
we evaluate the products using Wick's rule. Each covariance pairing is
computed according to the index-sharing pattern of the two paired overlap
entries. The local Wick contractions may produce background overlaps,
such as $\Omega_{jm}^{bb}$. After performing the index sums with the
normalization $C_{k,d}$, the remaining background moments are evaluated
by the same Gaussian moment rule, while diagonal moments use the
norm-concentration approximation. This closes the retained contractions
in terms of $Q_1$ and $Q_2$.
Here the closure is understood to hold at the leading
order of these normalized sums; this does not follow from finite-dimensional
Gaussianity alone. The relevant contractions for $Q_2$ are evaluated
explicitly in \cref{Sec:contractions}.

\section{Macroscopic flow of \texorpdfstring{$Q_1$}{Q1}}
\label{Sec:flow_Q_1}
\subsection{Expectations of triple products of Hermite polynomials}
\label{Sec:exp_triple}

We first derive the dynamics of $Q_1$ without spherical normalization. Substituting the online SGD step \cref{eq:sgd} into the definition of $Q_1$ gives
\begin{equation*}
    \begin{aligned}
          Q_1^{01}(t+1)  =  Q_1^{01}(t) + \frac{\eta}{kd}\sum_{\alpha=1}^d \sum_{i, j=1}^k W_{i \alpha}^0 g_{j \alpha} (t) ,    \qquad  Q_1^{11}(t+1)   =  Q_1^{11}(t) + \frac{2\eta}{kd}\sum_{\alpha=1}^d \sum_{i, j=1}^k  W_{i\alpha}(t)   g_{j\alpha}(t) .
    \end{aligned}
\end{equation*}
In the dynamics of $Q_1^{11}$, we omit the term
$\frac{\eta^2}{kd}\sum_{\alpha}\big(\sum_j g_{j\alpha}\big)^2$,
which acts as effective SGD noise~\cite{Veiga_ODE}.
This term is $O(\eta^2/d)$, but its contribution vanishes under
the learning-rate scaling specified in the main text.
For the same reason, we also omit it from \cref{eq:Q_p}.

For a given new training sample $(\bx_t, y^0_t)$, the gradient components $g_{j\alpha}(t)$ with respect to the student weights are given by:
\begin{equation*}
    \begin{aligned}
        g_{j\alpha}(t) & = -\frac{\partial\, (y^0_t - y_t)^2}{\partial W_{j\alpha}}    = 2(y^0_t - y_t) \frac{\partial y_t}{\partial W_{j\alpha}},
    \end{aligned}
\end{equation*}
where the teacher and student network outputs are expanded in the Hermite polynomial basis as:
\begin{equation*}
    \begin{aligned}
        y^0_t  = \frac{\mathbf{1}^\intercal}{\sqrt{k}} \sigma\Big( \frac{\bW^0 \bx_t}{\sqrt{d}} \Big) = \frac{1}{\sqrt{k}} \sum_{i=1}^k \sum_{\ell=0}^\infty \frac{\mu_\ell}{\ell!} \mathrm{He}_\ell(h_i^0), \qquad y_t    = \frac{\mathbf{1}^\intercal}{\sqrt{k}} \sigma\Big( \frac{\bW \bx_t}{\sqrt{d}} \Big) = \frac{1}{\sqrt{k}} \sum_{i=1}^k \sum_{\ell=0}^\infty \frac{\mu_\ell}{\ell!} \mathrm{He}_\ell(h_i),
    \end{aligned}
\end{equation*}
with pre-activations defined as $h_i^0 = [\bW^0 \bx_t]_i / \sqrt{d}$ and $h_i = [\bW \bx_t]_i / \sqrt{d}$. Hermite polynomials are orthogonal under
the Gaussian measure, and $\partial_x \mathrm{He}_n(x) = n \mathrm{He}_{n-1}(x)$ gives
\begin{equation*}
    \begin{aligned}
        \frac{\partial y_t}{\partial W_{j\alpha}} & = \frac{1}{\sqrt{k}} \sum_{i=1}^k \sum_{\ell=0}^\infty \frac{\mu_\ell}{\ell!} \frac{\partial}{\partial W_{j\alpha}} \mathrm{He}_\ell\Big( \frac{1}{\sqrt{d}}\sum_{\nu=1}^d W_{i\nu}x_{\nu} \Big)
        = \frac{x_\alpha}{\sqrt{dk}} \sum_{\ell=0}^\infty \frac{\mu_{\ell+1}}{\ell!} \mathrm{He}_\ell(h_j).
    \end{aligned}
\end{equation*}
Averaging over the training data then gives the update for $Q_1^{01}$:
\begin{equation*}
    \begin{aligned}
         \mathbb{E}_{\mathcal{D}_{t+1}}[Q_1^{01}(t+1)] ={}& \mathbb{E}_{\mathcal{D}_t}[Q_1^{01}(t)] + \frac{2\eta}{k^2d} \mathbb{E}_{\mathcal{D}_{t+1}}\Big\{ \Big( \sum_{i=1}^k h_i^0 \Big) \Big[ \sum_{j=1}^k \sum_{\ell=0}^\infty \frac{\mu_\ell}{\ell!} \big(\mathrm{He}_\ell(h_j^0) - \mathrm{He}_\ell(h_j)\big) \Big]  \sum_{m=1}^k \sum_{q=0}^\infty \frac{\mu_{q+1}}{q!} \mathrm{He}_{q}(h_m)  \Big\}.
    \end{aligned}
\end{equation*}

The order parameters $Q$ are empirical statistics of the overlap tensor $\bOmega$, which is itself determined by the weight matrix $\bW$. Since $\bW(t)$ is generated by online SGD, it depends on the previously observed samples $\mathcal{D}_t \coloneq  \{\bx_1, \dots, \bx_{t-1}\}$. This dependence is the source of randomness in the order parameters, so we take expectations over both the training history and the fresh sample used in the next SGD update. To evaluate the update of $Q_1^{01}$, we first isolate the input expectations of the following Hermite triplets:
\begin{equation*}
    \begin{aligned}
        I_1^{ijm}(\ell_2, \ell_3) \coloneq  \mathbb{E}_{\bx_t}\Big[ h_i^0 \mathrm{He}_{\ell_2}(h_j^0) \mathrm{He}_{\ell_3}(h_m) \Big], \qquad I_2^{ijm}(\ell_2, \ell_3) \coloneq  \mathbb{E}_{\bx_t}\Big[ h_i^0 \mathrm{He}_{\ell_2}(h_j) \mathrm{He}_{\ell_3}(h_m) \Big].
    \end{aligned}
\end{equation*}
Accordingly, we define the corresponding summations across modes and neuron indices as:
\begin{equation*}
    \begin{aligned}
        E_1 & \coloneq  \frac{1}{k^2} \sum_{i,j,m=1}^k \sum_{\ell_2, \ell_3 = 0}^\infty \frac{\mu_{\ell_2} \mu_{\ell_3+1}}{\ell_2! \ell_3!} I_1^{ijm}(\ell_2, \ell_3), \qquad E_2 & \coloneq  \frac{1}{k^2} \sum_{i,j,m=1}^k \sum_{\ell_2, \ell_3 = 0}^\infty \frac{\mu_{\ell_2} \mu_{\ell_3+1}}{\ell_2! \ell_3!} I_2^{ijm}(\ell_2, \ell_3).
    \end{aligned}
\end{equation*}
Using these definitions, the update rule simplifies to
\begin{equation}
        \mathbb{E}_{\mathcal{D}_{t+1}}[Q_1^{01}(t+1)] = \mathbb{E}_{\mathcal{D}_t}[Q_1^{01}(t)] + \frac{2\eta}{d} \mathbb{E}_{\mathcal{D}_t}[E_1 - E_2].\label{eq:Q1_sim}
\end{equation}
As discussed in the main text, the terms governed by $E_1$ (capturing the overlap between $h_j^0$ and $h_m$) act as the attractive force, whereas the terms in $E_2$ (overlap between $h_j$ and $h_m$) represent self-repulsion. We now express these terms through the order parameters at time $t$.

Applying Stein's lemma or, more directly, Lemma 12 from \citet{ricci2025}, we find that in the extensive-width limit these triplet expectations evaluate to:
    \begin{equation*}
        \begin{aligned}
            I_1^{ijm}={} & \sum_{\lambda=0}^1\ell_2! \ell_3!\,\frac{(\Omega^{00}_{ij})^{\lambda}(\Omega^{01}_{jm})^{\ell_3-1+\lambda}(\Omega^{01}_{im}-\Omega^{01}_{jm}\Omega^{00}_{ij})^{1-\lambda}}{
            \big(\frac{\ell_2-\ell_3+1}{2}\big)!
                \big(\frac{\ell_2+\ell_3-1}{2}\big)!
            \big(\frac{\ell_3-1+2\lambda-\ell_2}{2}\big)!}, \\
            I_2^{ijm}
            ={} & \sum_{\lambda=0}^1
            \ell_2! \ell_3!\,
            \frac{(\Omega^{01}_{ij})^{\lambda}
                (\Omega^{11}_{jm})^{\ell_3-1+\lambda}
                (\Omega^{01}_{im}-\Omega^{11}_{jm}\Omega^{01}_{ij})^{1-\lambda}}{\big(\frac{\ell_2-\ell_3+1}{2}\big)!
                \big(\frac{\ell_2+\ell_3-1}{2}\big)!
                \big(\frac{\ell_3-1+2\lambda-\ell_2}{2}\big)!}.
        \end{aligned}
    \end{equation*}
where terms with a negative or half-integer factorial argument in the denominator vanish. This imposes selection rules on the Hermite indices. Expanding $I_1^{ijm}$ explicitly yields:
\begin{equation*}
    \begin{aligned}
        I_1^{ijm}(\ell_3-1,\ell_3)
         & =
        \ell_3!(\Omega^{01}_{jm})^{\ell_3-1}\Omega^{01}_{im} \quad \text{for} \quad \ell_3\geq1,  \qquad I_1^{ijm}(\ell_3+1,\ell_3)
          =
        (\ell_3+1)!(\Omega^{01}_{jm})^{\ell_3}\Omega^{00}_{ij}
    \quad \text{for} \quad \ell_3\geq0 .
    \end{aligned}
\end{equation*}
Applying an analogous reduction to $I_2^{ijm}$ and substituting these back into the expressions for $E_1$ and $E_2$, we obtain the macroscopic equations:
    \begin{equation}
        \begin{aligned}
            E_1
            ={} & \frac{1}{k^2} \sum_{\ell=0}^\infty
            \Big[
                \frac{\mu_\ell \mu_{\ell+2}}{\ell!}
                \sum_{i,j,m=1}^k (\Omega^{01}_{jm})^\ell \Omega^{01}_{im}
                + \frac{\mu_{\ell+1}^2}{\ell!}
                \sum_{i,j,m=1}^k (\Omega^{01}_{jm})^\ell \Omega^{00}_{ij}
            \Big],                                  \\
            E_2
            ={} & \frac{1}{k^2} \sum_{\ell=0}^\infty
            \Big[
                \frac{\mu_\ell \mu_{\ell+2}}{\ell!}
                \sum_{i,j,m=1}^k (\Omega^{11}_{jm})^\ell \Omega^{01}_{im}
                + \frac{\mu_{\ell+1}^2}{\ell!}
                \sum_{i,j,m=1}^k (\Omega^{11}_{jm})^\ell \Omega^{01}_{ij}
                \Big].
        \end{aligned}
        \label{eq:E1E2}
    \end{equation}

\subsection{Scaling analysis of the overlap tensor}

To isolate the leading-order contributions to $Q_1$ in the extensive-width limit, we analyze the finite-size scaling of the tensor contractions involving $\Omega_{ij}$ in \cref{eq:E1E2}. For clarity, we temporarily suppress the teacher--student superscripts, such as $ab$, and use the teacher--teacher overlaps as a representative example. The same scaling estimates apply to the cross-overlap and student--student tensors, up to changes in the corresponding macroscopic moments. Since $k\propto d$, we express all asymptotic orders in powers of $d$. At initialization, the entries of the overlap tensor exhibit a diagonal--off-diagonal scale separation. For $i\ne j$, the off-diagonal overlaps verify $\mathbb{E}[\Omega_{ij}]=0,
    \mathrm{Var}(\Omega_{ij})=1/d$, and therefore have typical size $\Omega_{ij}=\Theta_{\mathbb P}(d^{-1/2})$. In contrast, the diagonal entries satisfy $\mathbb{E}[\Omega_{ii}]=1$ and remain $O(1)$, so that the normalized trace obeys $k^{-1}\sum_i\Omega_{ii}=\Theta(1)$. Similarly, for $i\ne j$, the squared overlaps scale as $\Omega_{ij}^2=\Theta_{\mathbb P}(d^{-1})$ with variance $O(d^{-2})$, whereas $\Omega_{ii}^2=\Theta(1)$ on the diagonal.

These entry-wise scaling estimates imply concentration of the macroscopic order parameters.  For example, the first two moments can be written as their deterministic limits plus finite-size fluctuations,
$$
    Q_1=\mathbb{E}[Q_1]+O_{\mathbb P}(d^{-1/2}),
    \qquad
    Q_2=\mathbb{E}[Q_2]+O_{\mathbb P}(d^{-1}).
$$
These fluctuations vanish in the thermodynamic limit and therefore do not affect the leading-order deterministic flow of the order parameters. This self-averaging property also justifies factorizing products of order parameters at leading order. Indeed, if $Q_a=\mathbb{E}[Q_a]+\delta Q_a$ and $Q_b=\mathbb{E}[Q_b]+\delta Q_b$, with $\delta Q_a,\delta Q_b$ vanishing in probability, then
$$
    \mathbb{E}[Q_aQ_b]
    =
    \mathbb{E}[Q_a]\mathbb{E}[Q_b]
    +
    o(1).
$$
Thus nonlinear couplings between macroscopic order parameters can be evaluated, to leading order, by replacing each order parameter with its deterministic expectation. Concentration makes the macroscopic dynamics deterministic and Markovian: the drift at step $t$ depends only on the current order parameters, as in \cref{eq:Q1_sim}, so there is no need to track the microscopic training history or average over past trajectories. Consequently, the online-SGD dynamics reduce to closed deterministic ODEs for a finite set of order parameters in the pre-specialization regime, rather than to history-dependent path-integral equations of the type that arise in dynamical mean-field theory. For notational simplicity hereafter, we identify the empirical order parameters with their deterministic large-system expectations:
\begin{equation*}
    \begin{aligned}
        Q_\ell(t) \equiv \mathbb{E}_{\mathcal{D}_{t}}[Q_\ell(t)] = \mathbb{E}_{\mathcal{D}_{t}}\Big[ \frac{1}{k}\sum_{i,j=1}^k (\Omega_{ij})^\ell \Big].
    \end{aligned}
\end{equation*}

To reduce \cref{eq:E1E2} to the $Q_1$ flow, we evaluate contractions of the form $\Omega_{ij}\Omega_{jm}^\ell$. \Cref{tab:Scaling} lists the scaling of each term; the last column includes the prefactor $1/d^3$ of \cref{eq:Q1_sim}.

\begin{table}[htbp]
    \centering
    \renewcommand{\arraystretch}{1.15}
    \begin{tabular}{cccccc}
        \hline\hline
        $\Omega_{ij}\Omega_{jm}^{\ell}$ & Index relation & Order & Sum of fluctuations & Multiplicity & Overall ($d^{-3}$) \\
        \hline
        $\ell = 0$  & $i\neq j$          & $O(d^{-1/2})$ & $d^2$ & $d$   & $O(d^{-3/2})$ \\
                    & $i = j$            & $O(1)$        & $1$   & $d^2$ & $O(d^{-1})$   \\
        \hline
        $\ell = 1$  & $i\neq j \neq m$   & $O(d^{-1})$   & $d^3$ & $1$   & $O(d^{-5/2})$ \\
                    & $i = j \neq m$     & $O(d^{-1/2})$ & $d^2$ & $1$   & $O(d^{-5/2})$ \\
                    & $i = m \neq j$     & $O(d^{-1})$   & $1$   & $d^2$ & $O(d^{-2})$   \\
                    & $i = j = m$        & $O(1)$        & $1$   & $d$   & $O(d^{-2})$   \\
        \hline
        $\ell = 2$  & $i\neq j \neq m$   & $O(d^{-3/2})$ & $d^3$ & $1$   & $O(d^{-3})$   \\
                    & $i\neq j = m$      & $O(d^{-1/2})$ & $d^2$ & $1$   & $O(d^{-5/2})$ \\
                    & $i = j \neq m$     & $O(d^{-1})$   & $1$   & $d^2$ & $O(d^{-2})$   \\
                    & $i = j = m$        & $O(1)$        & $1$   & $d$   & $O(d^{-2})$   \\
        \hline
        $\ell\geq3$ & $\cdots$           &               &       &       &               \\
        \hline\hline
    \end{tabular}
    \caption{Scaling of all the combinations of the form $\Omega_{ij}\Omega_{jm}^{\ell}$ for $\ell=0,1,2$ and $\ell\geq3$. The ``Index relation'' column lists the index configurations, ``Order'' gives the typical size of a single term, ``Sum of fluctuations'' the accumulated standard deviation over all index combinations, and ``Multiplicity'' the number of equivalent terms. The ``Overall'' column includes the normalization $1/(k^2d)$, which scales as $d^{-3}$ at fixed $k/d$; the learning-rate factor is omitted here. The centered-background estimates in this table do not include coherent contributions from nonzero overlap means during training, which are retained explicitly in the contractions below.}
    \label{tab:Scaling}
\end{table}

\begin{figure}[t]
    \centering
    \includegraphics[width=0.5\linewidth]{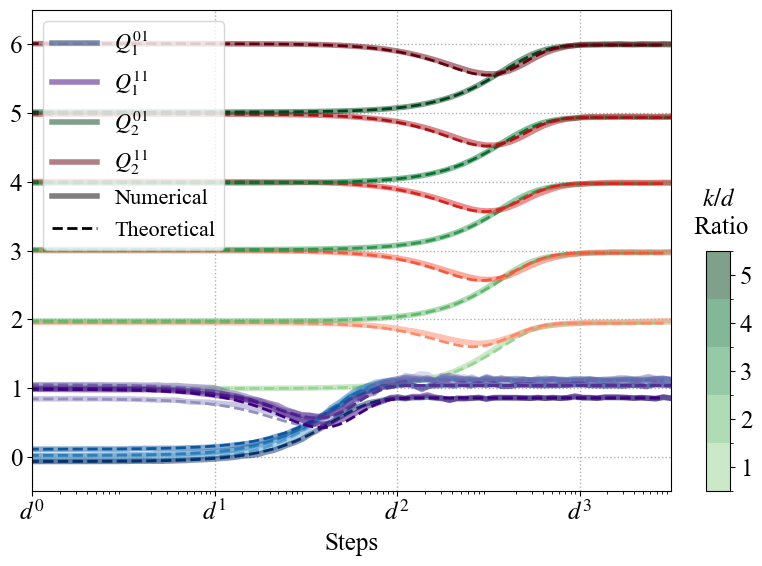}
    \caption{\label{fig:relu_dyn_kd} Macroscopic first- and second-order
    dynamics in a ReLU network. The input dimension is $d=100$, while the width
    $k$ is varied. Darker curves correspond to larger width-to-dimension ratios.
    Solid curves show online-SGD simulations, while dashed curves show the
    closed dynamical predictions from \cref{eq:q1_dyn,eq:q2_dyn}. For each run,
    the theoretical curve is initialized with the empirical order parameters of
    the corresponding seed. Thus the comparison tests the deterministic
    macroscopic flow conditioned on the initial state, rather than an
    ensemble-averaged trajectory.}
\end{figure}

Because $Q_1$ saturates on an $O(d/\eta)$ timescale, drift terms of order $O(\eta/d^{2})$ or smaller cannot accumulate to an $O(1)$ effect. Hence only the $\ell = 0$ sector contributes to the leading-order dynamics of $Q_1$.  According to \cref{eq:E1E2}, we have
\begin{equation*}
    \begin{aligned}
        E_1
        ={} & \frac{1}{k^2}
        \Big[
            \mu_0\mu_2\sum_{i,j,m=1}^k\Omega^{01}_{im}
            +\mu_1^2\sum_{i,j,m=1}^k\Omega^{00}_{ij}
        \Big]
        = \mu_0\mu_2 Q_1^{01}+\mu_1^2 Q_1^{00}, \\
        E_2
        ={} & \frac{1}{k^2}
        \Big[
            \mu_0\mu_2\sum_{i,j,m=1}^k\Omega^{01}_{im}
            +\mu_1^2\sum_{i,j,m=1}^k\Omega^{01}_{ij}
        \Big]
        = \mu_0\mu_2 Q_1^{01}+\mu_1^2 Q_1^{01}.
    \end{aligned}
\end{equation*}
Substituting into \cref{eq:Q1_sim} and proceeding similarly for $Q_1^{11}$ gives \cref{eq:q1_dyn} in the main text. \Cref{fig:relu_dyn_kd} compares the theoretical predictions with numerical simulations for several values of the width-to-dimension ratio $k/d$.

\section{Unclosed flow of general \texorpdfstring{$Q_p$}{Qp}}
\label{Sec:Overlap_Flow}
\subsection{Spherical normalization}

The derivation above omitted spherical normalization for clarity; including it does not change the leading-order dynamics of $Q_1$. Letting $\widetilde{\bW}(t+1) = \bW(t) + \eta\, \bg(t)$, the normalization step in \cref{eq:sgd} introduces the normalization factor raised to a power $n$, whose expansion in $\eta$ is
\begin{equation*}
    \begin{aligned}
         & \Big( \frac{\sqrt{kd}}{\|\widetilde{\bW}(t+1)\|_{\rm F}} \Big)^{n}
        =
        \Big[
            \frac{1}{kd}
            \sum_{i=1}^k \sum_{\alpha=1}^d
            \bigl(W_{i\alpha}(t)+\eta g_{i\alpha}\bigr)^2
        \Big]^{-n/2}   =
        1-\frac{n\eta}{kd}\sum_{i=1}^k\sum_{\alpha=1}^d W_{i\alpha}g_{i\alpha}
        -\frac{n\eta^2}{2kd}\sum_{i=1}^k\sum_{\alpha=1}^d g_{i\alpha}^2
        +O(\eta^3).\\
    \end{aligned}
\end{equation*}
Note that the $O(\eta^2)$ cross-term proportional to $(\sum_{i,\alpha} W_{i\alpha}g_{i\alpha})^2$ is smaller by a factor $1/d$ and hence subleading in the extensive-width limit. Focusing on the leading normalization term, we define
\begin{equation*}
    \begin{aligned}
        \frac{n\eta}{kd}\sum_{i=1}^k \sum_{\alpha=1}^d W_{i\alpha}g_{i\alpha} \coloneq  \frac{2n\eta}{d}(E^{\rm ret}_1 - E^{\rm ret}_2).
    \end{aligned}
\end{equation*}
As in the calculation of $Q_1$ without normalization, we average these terms over the data, which gives
        \begin{align*}
            \mathbb{E}_{\mathcal{D}_{t+1}}[E^{\rm ret}_1]
            &=  \frac{1}{k^2}\sum_{\ell=0}^{\infty}
            \mathbb{E}_{\bx_{t}}\Big[
                \frac{\mu_{\ell}\mu_{\ell+2}}{\ell!}
                \sum_{i,j=1}^k(\Omega^{01}_{ij})^{\ell}\Omega^{11}_{jj}
                +\frac{\mu_{\ell+1}^2}{\ell!}
                \sum_{i,j=1}^k(\Omega^{01}_{ij})^{\ell+1}
            \Big]                                    
            =  \frac{1}{k}\sum_{\ell=0}^{\infty}
            \Big[
                \frac{\mu_{\ell}\mu_{\ell+2}}{\ell!}Q^{01}_{\ell}
                +\frac{\mu_{\ell+1}^2}{\ell!}Q^{01}_{\ell+1}
            \Big],                                  \\
            \mathbb{E}_{\mathcal{D}_{t+1}}[E^{\rm ret}_2]
            &=\frac{1}{k^2}\sum_{\ell=0}^{\infty}
            \mathbb{E}_{\bx_{t}}\Big[
                \frac{\mu_{\ell}\mu_{\ell+2}}{\ell!}
                \sum_{i,j=1}^k(\Omega^{11}_{ij})^{\ell}\Omega^{11}_{jj}
                +\frac{\mu_{\ell+1}^2}{\ell!}
                \sum_{i,j=1}^k(\Omega^{11}_{ij})^{\ell+1}
            \Big]                                    
            = \frac{1}{k}\sum_{\ell=0}^{\infty}
            \Big[
                \frac{\mu_{\ell}\mu_{\ell+2}}{\ell!}Q^{11}_{\ell}
                +\frac{\mu_{\ell+1}^2}{\ell!}Q^{11}_{\ell+1}
                \Big].
    \end{align*}
The averaged normalization term is
\begin{equation*}
\begin{aligned}
     & \frac{n\eta}{kd}\,
        \mathbb{E}_{\mathcal{D}_{t+1}}\Big[
            \sum_{i=1}^k \sum_{\alpha=1}^d W_{i\alpha}g_{i\alpha}
            \Big]
            =
        \frac{2n\eta}{kd}\sum_{\ell=0}^{\infty}
        \Big[
            \frac{\mu_{\ell}\mu_{\ell+2}}{\ell!}
            (Q^{01}_{\ell}-Q^{11}_{\ell})
            +\frac{\mu_{\ell+1}^2}{\ell!}
            (Q^{01}_{\ell+1}-Q^{11}_{\ell+1})
            \Big].
\end{aligned}
\end{equation*}
This term is $O(\eta/d^2)$ and therefore does not affect $Q_1$ on the timescale $d/\eta$. Thus, in the linear-adaptation regime, the leading $Q_1$ dynamics are the same for unconstrained Euclidean SGD, Riemannian SGD, and explicit spherical normalization.

The $O(\eta^2)$ term of the expansion above contributes a correction of order
$\eta^2/d^2$, of the same order as the SGD-noise terms omitted in
\cref{eq:Q_p}. Both are controlled by the learning-rate condition given in the main text.

\subsection{Flow of \texorpdfstring{$Q_p$}{Qp}}
\label{Sec:Q_p_tensorial}

We now address the timescales up to specialization. Unlike the first-order phase, the dynamics of the higher-order parameters $Q_p$ ($p > 1$) contain tensor couplings, which do not reduce to functions of the $Q_p$. Recall the explicit trace definition of $Q_p^{ab}$:
\begin{equation*}
    Q_p^{ab}
    =
    \frac{1}{k d^p}
    \sum_{\alpha_1,\dots,\alpha_p=1}^d
    \sum_{i,j=1}^k
    \prod_{q=1}^{p}W_{i\alpha_q}^{a}W_{j\alpha_q}^{b}.
\end{equation*}
From $t=\Theta(d^2/\eta)$ to the onset of specialization, keeping the $O(\eta)$ contributions and averaging over the fresh sample gives the discrete-time updates:
    \begin{equation*}
        \begin{aligned}
              Q_p^{01}(t+1)
            &={}  \mathbb{E}_{\mathcal{D}_{t+1}}\Big[
                Q_p^{01}(t)
                \Big(
                1-\frac{p\eta}{kd}\sum_{i=1}^k\sum_{\alpha=1}^d W_{i\alpha}g_{i\alpha}
                \Big)
                +\frac{p\eta}{kd^p}
                \sum_{\alpha_1,\dots,\alpha_p=1}^d \sum_{i,m=1}^k
                W^0_{i\alpha_1}\cdots W^0_{i\alpha_p}
                W_{m\alpha_1}\cdots W_{m\alpha_{p-1}}
                g_{m\alpha_p}
            \Big]              \\
             & =  Q_p^{01}(t)
            -\frac{2p\eta}{kd} Q_p^{01}(t)
            \sum_{\ell=0}^\infty
            \Big[
                \frac{\mu_\ell\mu_{\ell+2}}{\ell!}(Q^{01}_\ell-Q^{11}_\ell)
                +\frac{\mu_{\ell+1}^2}{\ell!}(Q^{01}_{\ell+1}-Q^{11}_{\ell+1})
            \Big]             \\
             & \quad
            +\frac{2p\eta}{k^2d}
            \mathbb{E}_{\mathcal{D}_{t+1}}\Big\{
            \Big[
                \sum_{j=1}^k\sum_{\ell=0}^\infty
                \frac{\mu_\ell}{\ell!}
                \bigl(\mathrm{He}_\ell(h_j^0)-\mathrm{He}_\ell(h_j)\bigr)
                \Big]
            \Big[
                \sum_{i,m=1}^k\sum_{\ell=0}^\infty
                \frac{\mu_{\ell+1}}{\ell!}
                h_i^0\mathrm{He}_\ell(h_m)(\Omega_{im}^{01})^{p-1}
                \Big]
            \Big\},            \\
              Q_p^{11}(t+1)
            &=  Q_p^{11}(t)
            -\frac{4p\eta}{kd} Q_p^{11}(t)
            \sum_{\ell=0}^\infty
            \Big[
                \frac{\mu_\ell\mu_{\ell+2}}{\ell!}(Q^{01}_\ell-Q^{11}_\ell)
                +\frac{\mu_{\ell+1}^2}{\ell!}(Q^{01}_{\ell+1}-Q^{11}_{\ell+1})
            \Big]             \\
             & \quad
            +\frac{4p\eta}{k^2d}
            \mathbb{E}_{\mathcal{D}_{t+1}}\Big\{
            \Big[
                \sum_{j=1}^k\sum_{\ell=0}^\infty
                \frac{\mu_\ell}{\ell!}
                \bigl(\mathrm{He}_\ell(h_j^0)-\mathrm{He}_\ell(h_j)\bigr)
                \Big]
            \Big[
                \sum_{i,m=1}^k\sum_{\ell=0}^\infty
                \frac{\mu_{\ell+1}}{\ell!}
                h_i\mathrm{He}_\ell(h_m)(\Omega_{im}^{11})^{p-1}
                \Big]
            \Big\}.
        \end{aligned}
    \end{equation*}

The higher-order dynamics retain the attraction--repulsion structure, now through tensor contractions: the drift in $Q_p$ splits into competing attractive and repulsive triplet contractions. Generalizing \cref{eq:E1E2}, we define the leading-order triplet contractions $E^{(p)}_1$ and $E^{(p)}_2$ driving $Q_p^{01}$:
    \begin{equation}
        \begin{aligned}
            \mathbb{E}_{\mathcal{D}_{t+1}}[E^{(p)}_1]
            ={} & \frac{1}{k^2}\sum_{\ell=0}^{\infty}
            \mathbb{E}_{\mathcal{D}_{t}}\Big[
                \frac{\mu_{\ell}\mu_{\ell+2}}{\ell!}
                \sum_{i,j,m=1}^k
                (\Omega^{01}_{jm})^{\ell}(\Omega^{01}_{im})^p
                +\frac{\mu_{\ell+1}^2}{\ell!}
                \sum_{i,j,m=1}^k
                (\Omega^{01}_{jm})^{\ell}\Omega^{00}_{ij}
                (\Omega^{01}_{im})^{p-1}
            \Big],                                   \\
            \mathbb{E}_{\mathcal{D}_{t+1}}[E^{(p)}_2]
            ={} & \frac{1}{k^2}\sum_{\ell=0}^{\infty}
            \mathbb{E}_{\mathcal{D}_{t}}\Big[
                \frac{\mu_{\ell}\mu_{\ell+2}}{\ell!}
                \sum_{i,j,m=1}^k
                (\Omega^{11}_{jm})^{\ell}(\Omega^{01}_{im})^p
                +\frac{\mu_{\ell+1}^2}{\ell!}
                \sum_{i,j,m=1}^k
                (\Omega^{11}_{jm})^{\ell}\Omega^{01}_{ij}
                (\Omega^{01}_{im})^{p-1}
                \Big].
        \end{aligned}
        \label{eq:P_E1_E2}
    \end{equation}
To close the dynamics of $Q_p$, these contractions must be expressed as polynomials in the macroscopic order parameters.

\section{Macroscopic flow of \texorpdfstring{$Q_2$}{Q2}}
\label{Sec:contractions}

\subsection{Three-point contractions}
\label{subsec:Three-Point}
We first consider products of two overlaps with three distinct neuron indices, specifically, the $\mu_{\ell}\mu_{\ell+2}$ terms within $\mathbb{E}[E^{(p)}_1]$ and $\mathbb{E}[E^{(p)}_2]$ of \cref{eq:P_E1_E2}, and the $\mu_{\ell+1}^2$ terms at $\ell = 0, p=2$. In the extensive-width limit, sums over independent input coordinates concentrate, so these products can be decoupled with Wick's theorem. We also note that this calculation provides the covariance structure used in the definition of Gaussian closure. To make the contraction structure transparent, we represent each hidden-neuron index by a vertex and each overlap $\Omega$ by an edge between two vertices; the teacher--student labels are absorbed into the superscripts of the corresponding vertices. Taking the trace $\Omega^{01}_{im}\Omega^{00}_{ij}$ as an illustrative low-order example:
    \begin{equation}
        \begin{aligned}
             \frac{1}{k^2}\begin{tikzpicture}[
                                    baseline={([yshift=-1ex] current bounding box.center)},
                                    semithick
                                ]
                                \coordinate (m^1) at (0, 0);
                                \coordinate (i^0) at (0.6, 0);
                                \coordinate (j^0) at (1.2, 0);

                                \draw (m^1) -- (i^0);
                                \draw (i^0) -- (j^0);

                                \filldraw (m^1) circle (1pt) node[above=0pt] {$\scriptstyle m^1$};
                                \filldraw (i^0) circle (1pt) node[above=0pt] {$\scriptstyle i^0$};
                                \filldraw (j^0) circle (1pt) node[above=0pt] {$\scriptstyle j^0$};
                            \end{tikzpicture} &\coloneq  \frac{1}{k^2}\sum_{i,j,m=1}^k \Omega^{01}_{im}\Omega^{00}_{ij} = \frac{1}{d^2k^2}\sum_{i,j,m=1}^k\sum_{\alpha,\beta=1}^d W_{i\alpha}^0W_{m\alpha}^1W_{i\beta}^0W^0_{j\beta}                                                                                                 \\
             & \approx  \frac{1}{d^2k^2}\sum_{i,j,m=1}^k\sum_{\alpha,\beta=1}^d \wick{ \c1 W_{i\alpha}^0 \c2 W_{m\alpha}^1\c1 W_{i\beta}^0 \c2 W^0_{j\beta} } \delta_{\alpha\beta} + \frac{1}{d^2k^3}\sum_{i,j,m,s=1}^k\sum_{\alpha\neq\beta}^d \wick{ \c1 W_{i\alpha}^0 \c1 W_{m\alpha}^1\c2 W_{s\beta}^0 \c2 W^0_{j\beta} } \\
             & =  \frac{1}{d}Q_1^{01} + \frac{1}{k}Q_1^{00}Q_1^{01}.
             \label{eq:three_point}
        \end{aligned}
    \end{equation}
In the second step, we split the sum over input coordinates into $\alpha=\beta$ and $\alpha\neq\beta$. For $\alpha=\beta$ we use $\frac1k\sum_i (W^0_{i\alpha})^2\approx 1$, which gives $Q_1^{01}/d$. For $\alpha\neq\beta$, Gaussian closure treats $W^0_{i\alpha}$ and $W^0_{i\beta}$ as independent, and the sum factorizes into $Q_1^{00}Q_1^{01}/k$; the auxiliary index $s$ makes this factorization explicit. The same procedure gives the remaining three-point contractions:
\begin{align*}
\frac{1}{k^2}\sum_{i,j,m=1}^k
\Omega^{01}_{im}\Omega^{01}_{ij}
&\approx
\frac{1}{d}Q_1^{11}
+\frac{1}{k}(Q_1^{01})^2,\\
\frac{1}{k^2}\sum_{i,j,m=1}^k
\Omega^{11}_{im}\Omega^{10}_{ij}
&\approx
\frac{1}{d}Q_1^{01}
+\frac{1}{k}Q_1^{11}Q_1^{01},\\
\frac{1}{k^2}\sum_{i,j,m=1}^k
\Omega^{11}_{im}\Omega^{11}_{ij}
&\approx
\frac{1}{d}Q_1^{11}
+\frac{1}{k}(Q_1^{11})^2,\\
\frac{1}{k^2}\sum_{i,j,m=1}^k
\Omega^{01}_{jm}(\Omega^{01}_{im})^2
&\approx
\frac{2}{k^2d}\sum_{i,j,m=1}^k
\Omega_{ij}^{00}\Omega_{im}^{01}
+\frac{1}{k}Q_1^{01}Q_2^{01}
=\frac{1}{k}Q_1^{01}Q_2^{01}
+O\Big(\frac{1}{d^2}\Big).
\end{align*}
For coupled trace components of the form $\frac{1}{k}\sum_{i,j,m=1}^k (\Omega^{01}_{jm})^{\ell}(\Omega^{01}_{im} )^p$ with $\ell > 1$ and $p>1$, closure requires multiple simultaneous index contractions. These terms are at most of order $O(1/d^2)$ and therefore do not contribute at leading order. The $\ell = 2, p=2$ case is evaluated below:
    \begin{equation*}
        \begin{aligned}
             \frac{1}{k^2}\begin{tikzpicture}[
                                    baseline={([yshift=-0.5ex]current bounding box.center)},
                                    semithick
                                ]

                                \coordinate (i^1) at (0, 0);
                                \coordinate (m^1) at (0.7, 0);
                                \coordinate (j^0) at (1.4, 0);

                                \draw (i^1) to[bend left=45] (m^1);
                                \draw (i^1) to[bend right=45] (m^1);
                                \draw (m^1) to[bend left=45] (j^0);
                                \draw (m^1) to[bend right=45] (j^0);

                                \filldraw (i^1) circle (1pt) node[left=1pt] {$\scriptstyle i^1$};
                                \filldraw (m^1) circle (1pt) node[above=2pt] {$\scriptstyle m^1$};
                                \filldraw (j^0) circle (1pt) node[right=1pt] {$\scriptstyle j^0$};
                            \end{tikzpicture} &\coloneq  \frac{1}{k^2}\sum_{i,j,m=1}^k (\Omega^{01}_{jm})^2(\Omega^{01}_{im} )^2                                                                                                                                                                                    \\
             & \approx \sum_{i,j,m=1}^k\sum_{\alpha,\beta,\nu,\kappa}^d \frac{2\delta_{\alpha\nu} \delta_{\beta\kappa}}{k^2d^4}\wick{ \c1 W_{j\alpha}^0 \c2 W_{m\alpha}^1\c3 W_{j\beta}^0 \c4W_{m\beta}^1\c1 W_{i\nu}^0 \c2 W_{m\nu}^1 \c3 W_{i\kappa}^0 \c4W_{m\kappa}^1}  + \frac{1}{k}(Q_2^{01})^2 \\
             & = \frac{1}{k}(Q_2^{01})^2 + \frac{2}{d^2} Q_2^{00} =  \frac{1}{k}(Q_2^{01})^2 +O\Big(\frac{1}{d^2}\Big).
        \end{aligned}
    \end{equation*}
Hence, for all $\ell > 1$ and $p > 1$, trace terms of the generalized form $\frac{1}{k}\sum_{i,j,m=1}^k(\Omega^{ab}_{jm})^{\ell}(\Omega^{cd}_{im})^p$ asymptotically decouple into $\frac{1}{k}Q_\ell^{ab}Q_p^{cd} + O(1/d^2)$. Consequently, for any arbitrary $\ell$, the contraction components scaled by $\mu_\ell\mu_{\ell+2}/\ell!$ can be analytically reduced to:
\begin{equation*}
    \frac{2p\eta}{kd}\sum_{\ell=0}^\infty \frac{\mu_{\ell}\mu_{\ell+2}}{\ell!}Q_p^{01}(Q_\ell^{01}- Q_\ell^{11}).
\end{equation*}
This drift term is canceled by the corresponding $\mu_\ell \mu_{\ell+2}/\ell!$ terms generated by spherical normalization. Thus, to leading order, the macroscopic dynamics are governed by terms proportional to $\mu_{\ell+1}^2$, i.e., by the loop contractions~of~$\bOmega$.

\subsection{Loop contractions of \texorpdfstring{$Q_2$}{Q2}}

Loop contractions cannot be decoupled with the pairwise approximations above. Instead, all relevant coincidences of input and neuron indices must be evaluated at the leading scaling order. Below, we list the approximations needed to close the dynamics of $Q_2$. Before the specialization, we do not need to treat the diagonal term of $\Omega^{01}_{ij}$ specially. Consider the contraction of the loop corresponding to $\ell = 1$, $p = 2$ as an example:
\begin{equation*}
    \begin{aligned}
         \frac{1}{k^2}
        \begin{tikzpicture}[
                baseline={([yshift=-0.5ex]current bounding box.center)},
                semithick
            ]
            \coordinate (i^0) at (0, 0.6);
            \coordinate (j^0) at (0, -0.6);
            \coordinate (m^1) at (1.0, 0);

            \filldraw (i^0) circle (1pt) node[above left=-2pt] {$\scriptstyle i^0$};
            \filldraw (j^0) circle (1pt) node[below left=-2pt] {$\scriptstyle j^0$};
            \filldraw (m^1) circle (1pt) node[right=-1pt] {$\scriptstyle m^1$};

            \draw (i^0) to[bend right=50] (j^0);
            \draw (i^0) to[bend left=25] (m^1);
            \draw (j^0) to[bend right=25] (m^1);
        \end{tikzpicture} \coloneq  \frac{1}{k^2}\sum_{i,j,m=1}^k \Omega^{01}_{jm} \Omega^{01}_{im} \Omega^{00}_{ij}                 = \frac{1}{k^2d^3}\sum_{i,j,m=1}^k\sum_{\alpha,\beta,\nu=1}^d W_{j\alpha}^0 W_{m\alpha}^1W_{i\beta}^0W_{m\beta}^1W_{i\nu}^0 W^0_{j\nu}.
    \end{aligned}
\end{equation*}
The expansion of this loop involves two classes of index coincidences: (1) coincidences of input indices, $\alpha = \beta$ and $\alpha = \nu$; and (2) coincidences of neuron indices, $i=j$, with distinct input indices ($\alpha \neq \beta \neq \nu$).

For the first class, the leading terms come from the Wick pairings
\begin{equation*}
    \begin{aligned}
        (\alpha = \beta)        & : &  & \wick{ \c1 W_{j\alpha}^0 \c2 W_{m\alpha}^1\c1 W_{i\beta}^0 \c2W_{m\beta}^1\c3 W_{i\nu}^0 \c3 W^0_{j\nu} } \\
        (\alpha = \nu)         & : &  & \wick{ \c1 W_{j\alpha}^0 \c3 W_{m\alpha}^1\c2 W_{i\beta}^0 \c2W_{m\beta}^1\c3 W_{i\nu}^0 \c1 W^0_{j\nu} } \\
        (\beta = \nu) & : & & \wick{ \c1 W_{j\alpha}^0 \c1 W_{m\alpha}^1\c2 W_{i\beta}^0 \c3 W_{m\beta}^1\c2 W_{i\nu}^0 \c3 W^0_{j\nu} } \\
        (\alpha = \beta = \nu) & : &  & \wick{ \c1 W_{j\alpha}^0 \c2 W_{m\alpha}^1\c3 W_{i\beta}^0 \c2W_{m\beta}^1\c3 W_{i\nu}^0 \c1 W^0_{j\nu} }
    \end{aligned}
\end{equation*}
The configuration $\alpha=\beta=\nu$ is counted in all three input-index coincidence classes and must therefore
be subtracted twice.

For the second class, when the teacher indices coincide ($i=j$), the weights pair without any further coincidence of input indices:
\begin{equation*}
    \begin{aligned}
        (i=j)                             & : &  & \wick{\c1 W_{j\alpha}^0 \c1 W_{m\alpha}^1}\wick{\c1 W_{j\beta}^0 \c1 W_{m\beta}^1} \wick{\c1 W_{j\nu}^0 \c1 W^0_{j\nu} } \\
        (i=j \text{ and } \alpha = \beta) & : &  & \wick{ \c1 W_{j\alpha}^0 \c2 W_{m\alpha}^1\c1 W_{j\beta}^0 \c2W_{m\beta}^1} \wick{\c1 W_{j\nu}^0 \c1 W^0_{j\nu} }.
    \end{aligned}
\end{equation*}
We omit configurations such as $i=m$ or $j=m$ with $\alpha\neq\beta\neq\nu$ and $i\neq j$, or $i=j$ with $\alpha=\nu$ (or $\beta=\nu$), as they do not contribute at leading order.

Collecting these terms gives
    \begin{align*}
             \frac{1}{k^2}\sum_{i,j,m=1}^k \Omega^{01}_{jm} \Omega^{01}_{im} \Omega^{00}_{ij}
             & \approx \frac{1}{k^2}\sum_{i,j,m=1}^k\Big[\frac{1}{d}(\Omega_{ij}^{00})^2 + \frac{1}{d}(\Omega_{im}^{01})^2 + \frac{1}{d}(\Omega_{jm}^{01})^2\Big] -\frac{2k}{d^2} + \frac{1}{k}\Big(Q_2^{01}-\frac{k}{d}\Big) \\
             & = \frac{1}{d} (Q_{2}^{00} + 2Q_{2}^{01}-2k/d) + \frac{1}{k}\Big(Q_2^{01}-\frac{k}{d}\Big) + O\Big(\frac{1}{d^2}\Big).
    \end{align*}
The same expansion for the remaining loops gives
    \begin{equation*}
        \begin{aligned}
            \frac{1}{k^2}\sum_{i,j,m=1}^k
            \Omega^{11}_{jm}\Omega^{01}_{im}\Omega^{01}_{ij}
             & \approx
            \frac{1}{d}(Q_2^{11}+2Q_2^{01}-2k/d)
            +\frac{1}{k}\Big(Q_2^{01}-\frac{k}{d}\Big), \\
            \frac{1}{k^2}\sum_{i,j,m=1}^k
            \Omega^{01}_{jm}\Omega^{11}_{im}\Omega^{10}_{ij}
             & \approx
            \frac{1}{d}(Q_2^{11}+2Q_2^{01}-2k/d)
            +\frac{1}{k}\Big(Q_2^{01}-\frac{k}{d}\Big), \\
            \frac{1}{k^2}\sum_{i,j,m=1}^k
            \Omega^{11}_{jm}\Omega^{11}_{im}\Omega^{11}_{ij}
             & \approx
            \frac{1}{d}(3Q_2^{11}-2k/d-2)
            +\frac{1}{k}(2Q_2^{11}-1).
        \end{aligned}
    \end{equation*}
Also, for $\ell>1, p=2$
    \begin{equation*}
        \begin{aligned}
            \frac{1}{k^2}\sum_{i,j,m=1}^k
            (\Omega^{01}_{jm})^{\ell}\Omega^{00}_{ij}\Omega^{01}_{im}
             & \approx
            \frac{1}{d}Q_{\ell+1}^{01}
            +\frac{1}{k}Q_{\ell+1}^{01},          \\
            \frac{1}{k^2}\sum_{i,j,m=1}^k
            (\Omega^{11}_{jm})^{\ell}\Omega^{01}_{ij}\Omega^{01}_{im}
             & \approx
            \frac{1}{d}Q_{\ell+1}^{11}
            +\frac{1}{k}\Big(Q_{2}^{01}-k/d\Big), \\
            \frac{1}{k^2}\sum_{i,j,m=1}^k
            (\Omega^{01}_{jm})^{\ell}\Omega^{10}_{ij}\Omega^{11}_{im}
             & \approx
            \frac{1}{d}Q_{\ell+1}^{01}
            +\frac{1}{k}Q_{\ell+1}^{01},          \\
            \frac{1}{k^2}\sum_{i,j,m=1}^k
            (\Omega^{11}_{jm})^{\ell}\Omega^{11}_{ij}\Omega^{11}_{im}
             & \approx
            \frac{1}{d}(Q_{\ell+1}^{11}-1)
            +\frac{1}{k}(2Q_{\ell+1}^{11}+Q_2^{11}-2).
        \end{aligned}
    \end{equation*}

Combining the three-point and loop contractions above gives the pre-specialization dynamics of $Q_2$, \cref{eq:q2_dyn}.
\begin{figure}[t]
    \centering
    \includegraphics[width=1\linewidth]{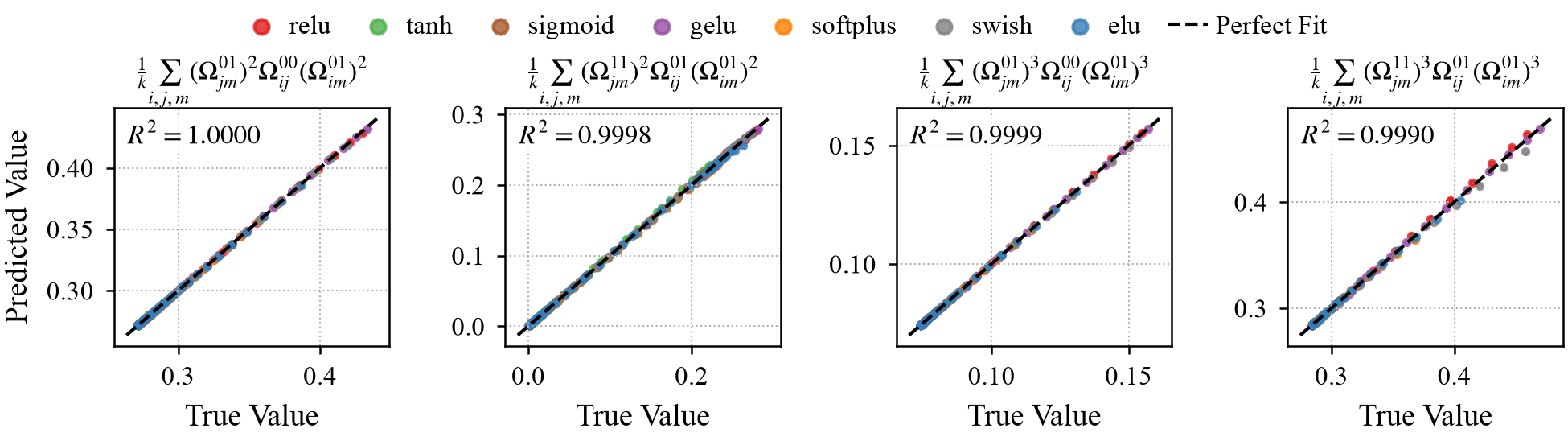}
    \caption{Fitted closure for high-order loop contractions. Data are collected from simulations with different activation functions in the unspecialized Gaussian-closure regime. The near-unit $R^2$ values show that a dictionary built from $Q_1^{ab}$ and $Q_2^{ab}$ captures the dominant finite-size loop contractions entering $Q_3$ and $Q_4$. This regression is used only to validate the Gaussian closure and is not part of the asymptotic scaling argument.}
    \label{fig:fit}
\end{figure}

\subsection{Higher-order loop contractions}

Beyond the $Q_2$ closure, the specialization channels involve higher-order
loop contractions such as
\begin{equation*}
    \begin{aligned}
& \frac{1}{k^2}
        \begin{tikzpicture}[
                baseline={([yshift=-0.5ex]current bounding box.center)},
                semithick
            ]
            \coordinate (i^0) at (0, 0.6);
            \coordinate (j^0) at (0, -0.6);
            \coordinate (m^1) at (1.0, 0);

            \filldraw (i^0) circle (1pt) node[above left=-2pt] {$\scriptstyle i^0$};
            \filldraw (j^0) circle (1pt) node[below left=-2pt] {$\scriptstyle j^0$};
            \filldraw (m^1) circle (1pt) node[right=-1pt] {$\scriptstyle m^1$};

            \draw (i^0) to[bend right=50] (j^0);
            \draw (i^0) to[bend left=-32] (m^1);
            \draw (i^0) to[bend left=32] (m^1);
            \draw (j^0) to[bend right=-32] (m^1);
            \draw (j^0) to[bend right=32] (m^1);
        \end{tikzpicture} \text{or} & \frac{1}{k^2}
        \begin{tikzpicture}[
                baseline={([yshift=-0.5ex]current bounding box.center)},
                semithick
            ]
            \coordinate (i^0) at (0, 0.6);
            \coordinate (j^0) at (0, -0.6);
            \coordinate (m^1) at (1.0, 0);

            \filldraw (i^0) circle (1pt) node[above left=-2pt] {$\scriptstyle i^0$};
            \filldraw (j^0) circle (1pt) node[below left=-2pt] {$\scriptstyle j^0$};
            \filldraw (m^1) circle (1pt) node[right=-1pt] {$\scriptstyle m^1$};

            \draw (i^0) to[bend right=50] (j^0);
            \draw (i^0) to[bend left=8] (m^1);
            \draw (i^0) to[bend left=-32] (m^1);
            \draw (i^0) to[bend left=32] (m^1);
            \draw (j^0) to[bend right=8] (m^1);
            \draw (j^0) to[bend right=-32] (m^1);
            \draw (j^0) to[bend right=42] (m^1);
        \end{tikzpicture}
    \end{aligned}
\end{equation*}
which contribute to the leading contractions of high order dynamics in \cref{eq:Q_p}. Each of these
loops can, in principle, be reduced by enumerating all index coincidences
and Wick pairings, as in the calculation of $Q_2$ above. However, the number
of Wick pairings grows factorially with the contraction order, making direct
enumeration impractical at high orders. One possible approach is to approximate each
high-order contraction by a polynomial in the low-order macroscopic
observables $Q_1^{ab}$ and $Q_2^{ab}$. We retain only monomials that preserve
the numbers of teacher and student factors in the original loop and fit
their coefficients to simulation data in the unspecialized regime.

The resulting fitted closures for the high-order contractions are shown in \cref{fig:fit}. 
The corresponding $R^2$ values are close to one, indicating that this
low-order dictionary captures the dominant loop structure of the Gaussian regime
at the simulated finite sizes. 
The main source of residual error is the
approximation $Q_{W_2}^{11}=1$. In finite dimension,
$Q_{W_2}^{11}=1+ o(1)$, and loops containing several student--student edges
multiply this correction several times. The error is therefore amplified for
higher-order loops, consistent with the gradual decrease of $R^2$ in the
panels with more student--student edges in \cref{fig:fit}. 

\section{The onset of specialization and its speed-up through quadratic shielding}%
\label{app:sec:specialization}

We define the \emph{onset of specialization} as the first time at which
$Q_{W_\ell}^{01}$ exceeds its value at the saturation of $Q_1$ and $Q_2$ by a
fixed positive amount independent of the system size. We call the interval
between this saturation and the onset of specialization the \emph{quadratic
plateau}. Here, we seek to estimate the timescale at which specialization sets
in,  Throughout, we work at fixed $k/d$ and use the unit-diagonal approximation
$\Omega_{ii}^{00}=\Omega_{jj}^{11}=1$.

\subsection{Effective equation of motion for teacher-student overlap}\label{app:E1}

We will obtain estimates for the onset of specialization by deriving an
effective equation of motion for a single teacher-student overlap, including the
leading quadratic SGD correction. First, we note that at the end of Gaussian closure, the student has learned a quadratic
approximation of the teacher, which amounts to matching its first three Hermite
modes. We hence assume that the first two modes of the generalization error in
\cref{eq:loss_decomp} are zero, i.e.\ $Q_p^{00}-2Q_p^{01}+Q_p^{11}=0$
for $p=1,2$, or more explicitly
\begin{align*}
Q_1^{00}-2Q_1^{01}+Q_1^{11}
&=\frac{1}{kd}\Big\|
\sum_i\mathbf w_i^1-\sum_i\mathbf w_i^0
\Big\|^2,\\
Q_2^{00}-2Q_2^{01}+Q_2^{11}
&=\frac{1}{kd^2}\big\|
\bW^{1\intercal}\bW^1-\bW^{0\intercal}\bW^0
\big\|_{\rm F}^2.
\end{align*}
At the end of Gaussian closure, we therefore have the following two conditions on the weights:
\begin{equation}
\sum_i\mathbf w_i^1=\sum_i\mathbf w_i^0,
\qquad
\bW^{1\intercal}\bW^1=\bW^{0\intercal}\bW^0.
\label{app:eq:low_order_matching_qw}
\end{equation}

The matching conditions \cref{app:eq:low_order_matching_qw} cancel the
population-drift terms proportional to $\mu_1^2$, $\mu_2^2$, $\mu_1\mu_3$, and
$\mu_2\mu_4$ in the master equation \cref{eq:Q_p}; the $\mu_0\mu_2$ term
vanishes identically. Since $r$ is the first active Hermite sector above
quadratic order, $\mu_t=0$ for $3\le t\le r-1$, so the remaining mixed terms
$\mu_\ell\mu_{\ell+2}$ with $\ell<r$ vanish as well. Under Gaussian closure, the
mixed terms with $\ell\ge r$ and the higher pure Hermite sectors are subleading
at the Gaussian overlap scale, for fixed coefficients. For example, the
$\mu_3\mu_5$ contribution for $r=3$ is suppressed by $O(d^{-1/2})$ relative to
the $\mu_3^2$ source. These cancellations and scale estimates motivate
considering an activation function $\sigma(z)$ that retains the learned
quadratic part and the first active higher Hermite sector. For concreteness, we
therefore consider the effective activation function
\begin{equation}
    \label{app:eq:effective-activation}
\sigma(z)=\mu_0+\mu_1z+\frac{\mu_2}{2}(z^2-1)
+\varepsilon\,\psi_r(z),
\qquad
\psi_r(z)=\frac{\mathrm{He}_r(z)}{\sqrt{r!}},
\end{equation}
which has specialization exponent $r\geq3$; $\psi_r(z)$ is normalized to have
unit norm w.r.t.\ the standard Gaussian measure, and
$\varepsilon=\mu_r/\sqrt{r!}$.

For a fresh Gaussian input $\bx$,
let $z_j=\mathbf w_j\cdot\bx/\sqrt d$, $u_i=\mathbf w_i^0\cdot\bx/\sqrt d$, and $e=f_0(\bx)-f_{\bW}(\bx)$
be the error between teacher and student, with $\bW\equiv\bW^1$. We normalise the
full weight matrix as in \cref{eq:sgd}, so that $\|\bW\|_F^2=kd$, retaining the
unit-diagonal approximation for the individual neuron norms. This gives the
normalized update
\begin{equation}
\bg_j=\frac{2e\,\sigma'(z_j)}{\sqrt{kd}}\bx,
\qquad
\mathbf w_j^+=\frac{\mathbf w_j+\eta\bg_j}{\sqrt{1+2\eta S+\eta^2V}},
\end{equation}
where
\begin{equation}
S=\frac{\langle\bW,\bg\rangle_F}{kd},
\qquad
V=\frac{\|\bg\|_F^2}{kd},
\qquad
H=\frac{\mathbf w_i^0\cdot\bg_j}{d}.
\end{equation}
For $\omega=\Omega_{ij}^{01}$, expanding the exact relation $\omega^+=(\omega+\eta
H)/\sqrt{1+2\eta S+\eta^2V}$ in $\eta$ gives
\begin{equation}
\label{eq:spec_second_order}
\Delta \omega
=\eta(H-\omega S)
+\eta^2\Big[-\frac12\omega V+\frac32\omega S^2-HS\Big]
+O(\eta^3).
\end{equation}
The linear term is the sample gradient projected onto the global Frobenius sphere. The term $-\eta^2\omega V/2$ describes the shrinkage induced by normalizing a step of nonzero squared length; its mean contains both deterministic and stochastic gradient contributions.

At the end of Gaussian closure, the student has learned a quadratic
approximation of the teacher so by  \cref{app:eq:low_order_matching_qw},the
error term is simply
\begin{equation}
e=\varepsilon R_r,
\qquad
R_r=\frac1{\sqrt k}\sum_a[\psi_r(u_a)-\psi_r(z_a)].
\end{equation}
Under the same delocalized Gaussian-plateau approximation,
$\mathbb ES^2,\mathbb E|HS|=O(\varepsilon^2/(kd^2))$, whereas $\mathbb
EV=\Theta(\varepsilon^2/(kd))$.  Relative to $\omega\mathbb EV$, the discarded terms
are therefore $O(d^{-1})$ and $O(d^{-1}|\omega|^{-1})$, respectively, and are
subleading for overlaps with $|\omega|\gtrsim d^{-1/2}$, including the leading
overlaps considered below.

To compute the mean drift $\mathbb E[H-\omega S]$, we note that $\mathbb{E} R_r u_i =
\mathbb{E} R_r u_i z_j = 0$ since $R_r$ is proportional to the $r$th Hermite
polynomial and $u_i, z_j$ give at most a degree-two factor. 
Therefore, the terms linear in $\varepsilon$ vanish and we find that
\begin{equation}
\mathbb E[H-\omega S]=\frac{2r\varepsilon^2}{kd}\mathcal F_{ij},
\end{equation}
where
\begin{equation}
\mathcal F_{ij} =\sum_a\Omega_{ia}^{00}(\Omega_{aj}^{01})^{r-1}
-\sum_a\Omega_{ia}^{01}(\Omega_{aj}^{11})^{r-1}-\omega(Q_r^{01}-Q_r^{11}).
\label{eq:spec_F}
\end{equation}
Thus the population drift is proportional to $\varepsilon^2$, even
though the sample gradient contains an order-$\varepsilon$ contribution.

For the quadratic correction, we use again Gaussian closure to obtain 
\begin{equation}
\mathbb EV
=\frac{4\varepsilon^2}{kd}
\mathbb E\Big[
R_r^2\frac{\|\bx\|^2}{d}
\frac1k\sum_j\sigma'(z_j)^2
\Big]
\approx\frac{4\varepsilon^2\mathcal{L}_ra_\sigma}{kd},
\end{equation}
where we have written $a_\sigma=\mathbb
E[\sigma'(Z)^2]=\mu_1^2+\mu_2^2+r\varepsilon^2$ and $\mathcal{L}_r=\mathbb E
R_r^2=\Theta(1)$. Combining the two equations, we obtain the equation of motion
\begin{equation}
\label{app:eq:specialization_eom}
\mathbb E[\Delta \omega\mid\bW]
\approx
\frac{2r\eta\varepsilon^2}{kd}\mathcal F_{ij}
-\frac{2\eta^2\varepsilon^2\mathcal{L}_ra_\sigma}{kd}\omega.
\end{equation}

\subsection{Estimating the timescale of specialization} \label{app:E2}

\begin{figure}[t]
    \centering
    \includegraphics[width=0.9\linewidth]{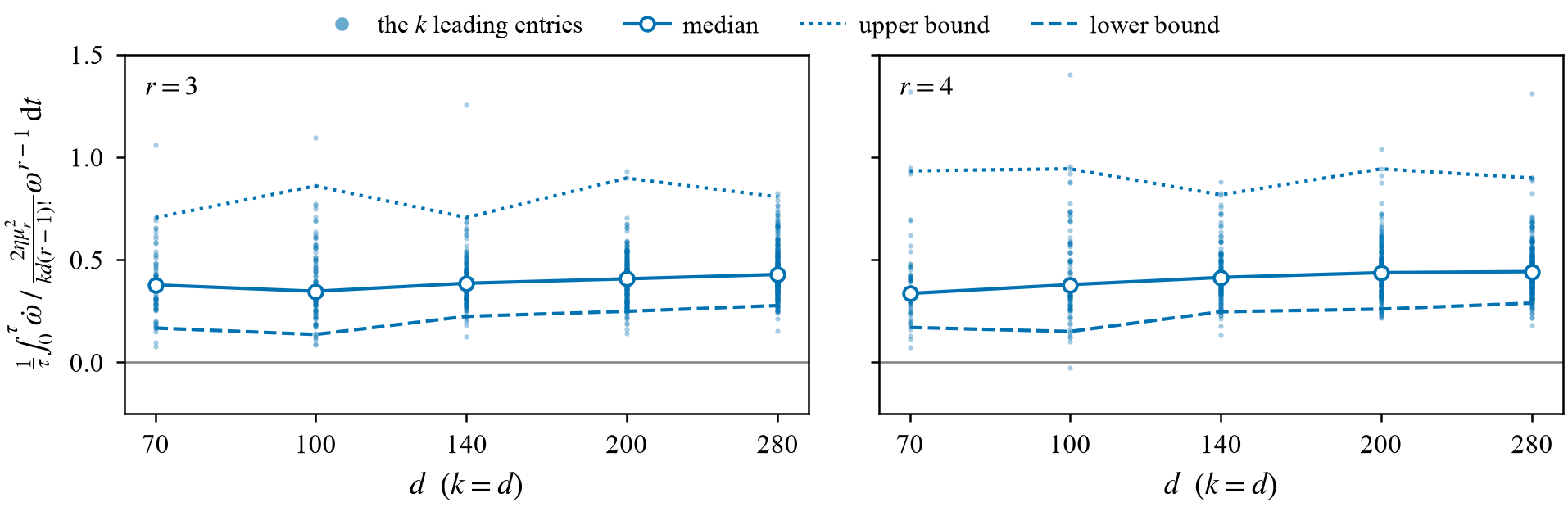}
    \caption{Numerical check of \cref{eq:g_average}. The  x-axis
shows the time-averaged ratio in \cref{eq:g_average}, and the
y-axis indicates the dimension $d\ (k=d)$. Each point is
computed over the interval from the saturation of $Q_1$ and
$Q_2$ to specialization onset. We define onset as the first
time $Q_{W_\ell}$ increases by $0.2$ above its plateau value,
tracking the largest-magnitude student--teacher overlap
for each student. The lower bound is estimated by the 5th percentile of the data, while the upper bound is the maximum after excluding the largest
$1/d$ fraction of values.  }
    \label{fig:bound_num}
\end{figure}

To estimate the growth driven by the population term, consider
$\mathcal F_{ij}$ in \cref{eq:spec_F}. The $a=i$ term of its first sum is
$\omega^{r-1}$. The $a=j$ term of its second sum cancels the diagonal part of
the normalization term, leaving the two off-diagonal sums and
$-\omega(Q_r^{01}-Q_{r,\mathrm{off}}^{11})$, where
$Q_{r,\mathrm{off}}^{11}=Q_r^{11}-1$ in the unit-diagonal approximation.
At $|\omega|\asymp d^{-1/2}$, the source $\omega^{r-1}$ has magnitude
$\Theta(d^{-(r-1)/2})$. Gaussian closure gives off-diagonal fluctuations
of order $O(d^{-(r-1)/2})$ and a remaining normalization term of order
$O(|\omega|d^{-(r-2)/2})$ for even $r$, smaller by $d^{-1/2}$ for odd $r$.
These estimates do not exclude cancellation of the source, so a positive
growth assumption is still needed to infer a specialization timescale.

For each student, let $\omega$ be the magnitude of its largest positive
overlap for odd $r$, or its largest absolute overlap for even $r$.
For the population dynamics, we assume that growth from the initial
plateau value $\omega_0$ to a fixed threshold
$\omega_{\mathrm{spec}}\in(0,1)$ in time $\tau$ satisfies
\begin{equation}
0<c_-\le\frac1\tau\int_0^{\tau}
\frac{\dot\omega}{\frac{2r\eta\varepsilon^2}{kd}\,\omega^{r-1}}
\,\mathrm dt\le c_+.
\label{eq:g_average}
\end{equation}
The constants are independent of $k$ at fixed $r$ and $k/d$. The lower
bound excludes cancellation or reversal of growth on average; extending
the bounds beyond the Gaussian regime is an additional assumption.
For collective onset, the lower bound is needed for a sufficiently large
fixed fraction of students, and the upper bound for all but a vanishing
fraction. A numerical check is provided in \cref{fig:bound_num}.

The initial condition is set at the beginning
of the plateau, after $Q_1$ and $Q_2$ have saturated. At this point,
Gaussian closure gives a mean $Q_1^{01}/k$ and a standard deviation
approximately equal to $\sqrt{Q_2^{01}/k}$ for each
$\Omega_{ij}^{01}$. The mean is negligible compared with the extreme
level. Applying the Gaussian extreme-value estimate to the $k$
teacher overlaps of each student therefore gives
\begin{equation*}
\omega_0\asymp
\sqrt{\frac{2Q_2^{01}\log k}{k}}
=
\Theta\Big(\sqrt{\frac{\log k}{k}}\Big),
\end{equation*}
where $Q_2^{01}=\Theta(1)$ on the initial plateau. At fixed $k/d$, this is also $\Theta(\sqrt{\log k/d})$.

\begin{figure}[t]
    \centering
    \includegraphics[width=0.9\linewidth]{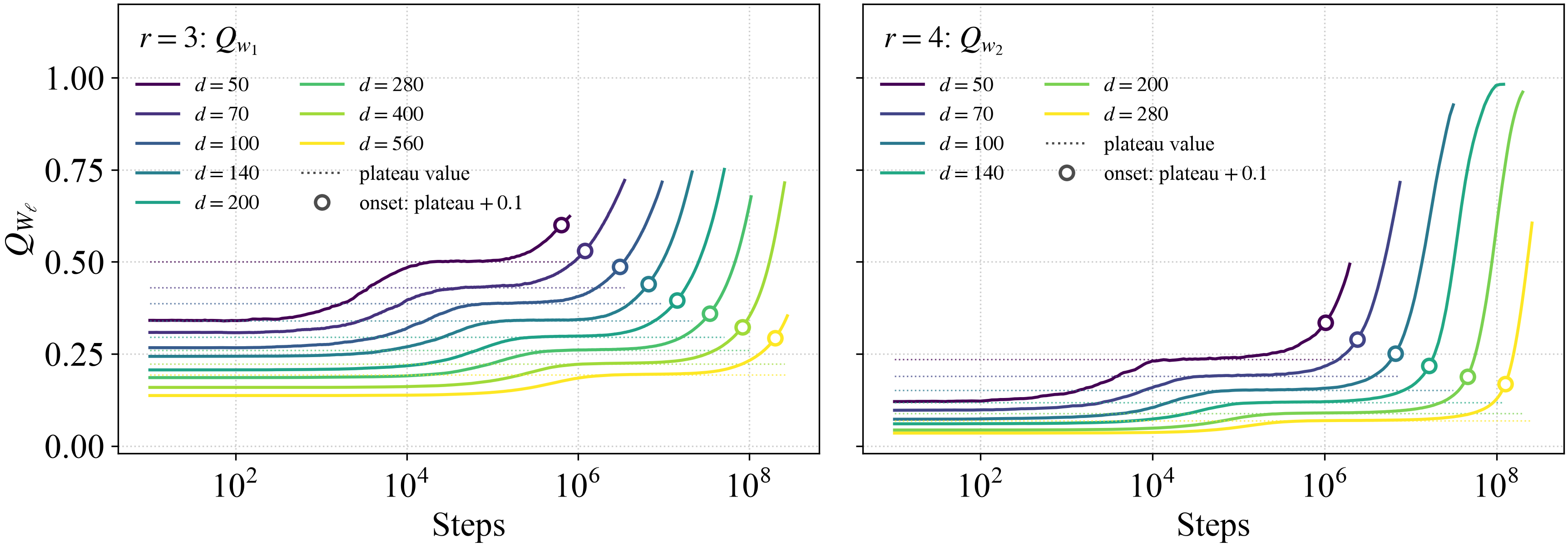}
    \caption{\label{fig:specialization_trajs} Evolution of $Q_{W_\ell}^{01}$ for different widths $k$, with
$d=k$, illustrating how the timescale data in
\cref{fig:timescale_exp-vs-theo} are collected.
We show $r=3$ and $r=4$ as examples.
Since the value of $Q_{W_\ell}^{01}$ at the beginning of
the plateau depends on $d$, we measure the first time it
exceeds this value by a fixed amount of $0.1$.
Alternatively, one could use a fixed absolute threshold,
such as $Q_{W_\ell}^{01}=0.6$, for all dimensions.
We expect the same asymptotic scaling, provided the subsequent
growth introduces no additional slow timescale.
Indeed, the reduced dynamics predict that growth between
two fixed positive overlap magnitudes takes $O(k^2/(\eta\varepsilon^2))$
time, which is asymptotically shorter than the plateau.
A higher threshold can nevertheless introduce stronger
finite-size effects at small $d$ and requires longer
simulations at large $d$.}
\end{figure}

Neglecting the term in $\eta^2$, \cref{eq:g_average} gives
\begin{equation}
\frac{\omega_0^{-(r-2)}-\omega_{\mathrm{spec}}^{-(r-2)}}{r-2}
=\int_0^\tau\frac{\dot\omega}{\omega^{r-1}}\,\mathrm dt
=\Theta\Big(\frac{2r\eta\varepsilon^2}{kd}\tau\Big).
\label{eq:reduced_dyn}
\end{equation}
With this initial condition, the initial-overlap term dominates, yielding
\begin{equation}
    \tau_{\mathrm{GF}}
=\Theta\Big(
\frac{k^{1+r/2}}
{\eta\varepsilon^2(\log k)^{(r-2)/2}}
\Big),
\end{equation}
recovering \cref{eq:tau_spec_gf} at fixed $\varepsilon$.
When a sufficiently large fixed fraction of students reach
$\omega_{\mathrm{spec}}$, we assume that a fixed positive fraction of their
leading overlaps can still form one-to-one teacher matches. Their growth
then produces a fixed increase in $Q_{W_\ell}^{01}$, with $\ell$ chosen even
when $r$ is even to avoid sign cancellation. Thus the overlap growth time
sets the collective specialization timescale. The predicted scaling is tested
in \cref{fig:timescale_exp-vs-theo}, with the measurement procedure described
in \cref{fig:specialization_trajs}.

Using the same effective growth estimate, balancing the two terms in
\cref{app:eq:specialization_eom} at $\omega_0$ gives the noise-limited learning-rate scale
\begin{equation}
\eta_{c}
\asymp
\frac{r}{\mathcal{L}_ra_\sigma}\omega_0^{r-2}.
\end{equation}
Choosing a sufficiently small fixed fraction of this scale gives
\begin{equation}
  \tau_{\rm spec}=\Theta\Big(
\frac{a_\sigma}{\varepsilon^2}
\frac{k^r}{(\log k)^{r-2}}
\Big).
\end{equation}

\subsection{Quadratic shielding speeds up specialization}%
\label{app:sec:quadratic-shielding}

So far, we have used low-order learning to simplify the residual error,
but have not incorporated its dynamical confinement of the weights.
This partial learning has an important impact on the learning dynamics even
though it is achieved without
teacher-student specialization. To see this, we now constrain the learning
dynamics to occur on the manifold that is defined by the two conditions of the
student weights that result from learning the quadratic part of the teacher,
namely the matching conditions \cref{app:eq:low_order_matching_qw}:
\begin{equation}
    \label{app:eq:gc-manifold}
\mathcal M=
\big\{
\bW:\mathbf1^\intercal\bW=\mathbf1^\intercal\bW^0,\ \bW^\intercal\bW=\bW^{0\intercal}\bW^0
\big\}.
\end{equation}
By the timescale separation between learning the quadratic part of the teacher
and specialization discussed in \cref{sec:observations}, we assume that
relaxation normal to this manifold is sufficiently fast to approximate the slow
dynamics of specialization by the projected gradient $\bg_{\mathcal{M}}=\Pi_{\bW}\bg$, where $\Pi_{\bW}$
is the orthogonal projection onto the tangent space $T_{\bW}\mathcal{M}$. This
effective-confinement assumption is additional to Gaussian closure; ordinary
globally normalized SGD does not enforce these constraints sample by sample.

To see how restricting the dynamics of the weights to the manifold affects the
SGD dynamics, we first note that every tangent perturbation $\bA$ satisfies
$\mathbf1^\intercal\bA=0$ and $\bW^\intercal\bA+\bA^\intercal\bW=0$, since the conditions of \cref{app:eq:gc-manifold} must hold to first order in $\bA$. Therefore, the
directional derivative of the quadratic part of the student network is
\emph{exactly} zero:
\begin{equation}
D f_{\le2,\bW}(\bx)[\bA]
=
\frac{\mu_1}{\sqrt{kd}}\mathbf1^\intercal\bA\bx
+\frac{\mu_2}{\sqrt k\,d}\bx^\intercal\bW^\intercal\bA\bx
=0.
\end{equation}
Writing $f_{r,\bW}(\bx)=k^{-1/2}\sum_j\psi_r(z_j)$, the low-degree sample
gradient is thus normal to $\mathcal M$, and
\begin{equation}
\bg_{\mathcal{M}}
=\Pi_{\bW}\big[
2\varepsilon R_r\nabla_{\bW} f_{\le2,\bW}
+2\varepsilon^2R_r\nabla_{\bW} f_{r,\bW}
\big]
=2\varepsilon^2R_r\Pi_{\bW}\nabla_{\bW} f_{r,\bW}.
\end{equation}
Unlike the cancellation in the population drift, this cancellation holds for
every input. It removes the gradient component linear in $\varepsilon$; the
projected drift retains its prefactor $\eta\varepsilon^2$.
For the effective confined dynamics, we assume that its leading-overlap
growth still satisfies the scaling in \cref{eq:g_average}. The norm of the
gradient is affected by the exact cancellation: writing
$\mathbf b=(b_j)_{j\le k}$ with $b_j=\sqrt r\,\psi_{r-1}(z_j)$, we obtain
\begin{equation}
\mathbb EV_{\mathcal{M}} \coloneq  \frac{\mathbb E\|\bg_{\mathcal{M}}\|_F^2}{kd}
=
\frac{4\varepsilon^4}{k^2d^2}
\mathbb E\big[
R_r^2\|\Pi_{\bW}(\mathbf b\bx^\intercal)\|_F^2
\big],
\end{equation}
where the factor $\varepsilon^4$ is exact. To estimate the dimension
scaling, we additionally assume
$\mathbb E[R_r^2\|\Pi_{\bW}(\mathbf b\bx^\intercal)\|_F^2]=\Theta(kd)$, so that
$\mathbb EV_{\mathcal{M}}=\Theta(\varepsilon^4/(kd))$.
Assuming that the effective confined dynamics have a growth-opposing
second-order correction of scale $\eta^2\omega\mathbb EV_{\mathcal{M}}$, balancing again drift and noise gives
the revised estimate of the optimal learning rate, and hence specialization
timescale,
\begin{equation}
\eta_{c,\mathcal M}
=\Theta\Big(
\frac{\omega_0^{r-2}}{\varepsilon^2}
\Big),
\qquad
\tau_{\mathrm{spec}}
=\Theta\Big(
\frac{k^r}{(\log k)^{r-2}}
\Big).
\end{equation}
Here again the time is evaluated at a sufficiently small fixed fraction of the
critical scale, in the regime where this is the active step-size restriction.
Compared with the naïve derivation, the admissible learning rate increases since
there is less noise, so the optimized learning time decreases by a factor of
order $a_\sigma/\varepsilon^2$. However, note that the dimension exponent is
unchanged. At a fixed learning rate, the cancellation instead enlarges the
regime in which the gradient-flow prediction remains valid, from $\eta
a_\sigma\ll\omega_0^{r-2}$ to $\eta\varepsilon^2\ll\omega_0^{r-2}$.

\subsection{Numerical verification of quadratic shielding predictions}

An important assumption in the analysis of quadratic shielding is that the fast
dynamics of $Q_1$ and $Q_2$ effectively confine SGD to the manifold
$\mathcal{M}$, \cref{app:eq:gc-manifold}, after the Gaussian plateau. We test
this effective confinement by measuring the diffusion of teacher--student
overlaps on the Gaussian plateau. After learning the quadratic part of the
teacher, but before specialization, the dynamics of teacher-student overlaps are
dominated by stochastic increments over lags that are sufficiently shorter than
the specialization time. From the equation of motion for an general overlap $\omega$ in
\cref{eq:spec_second_order}, we see that the leading contribution to the
variance is
\begin{equation}
    \var (\Delta \omega) = \eta^2 \var (H - \omega S) + \mathcal{O}(\eta^3),
\end{equation}
always at fixed weights. On the plateau, where a typical $\omega$ has size
$d^{-1/2}$, $H$ and $S$ both have root-mean-square size of order
$\varepsilon/(d\sqrt{k})$. Thus $\omega S$ is smaller than $H$ by a factor of order
$d^{-1/2}$. To leading order, the overlap noise therefore comes from $\eta H$.
Under the same assumptions as in the derivation of, we find 
\begin{equation}
v\coloneq  \var (\Delta \omega) \simeq
\frac{4\eta^2\varepsilon^2}{kd^2}
\mathbb E[R_r^2u_i^2\sigma'(z_j)^2]
\simeq
\frac{4\eta^2\varepsilon^2\mathcal{L}_ra_\sigma}{kd^2}.
\label{app:eq:overlap_noise_full}
\end{equation}

For the projected dynamics on $\mathcal M$, the sample-wise cancellation derived
above instead gives $\bg_{\mathcal{M}}=2\varepsilon^2R_r\Pi_{\bW}\nabla_{\bW} f_{r,\bW}$ and hence
\begin{equation}
v_{\mathcal M}\coloneq \var(\Delta \omega_{\mathcal{M}})
\simeq
\frac{4 \eta^2\varepsilon^4}{kd^2} \mathbb{E} \big[
R_r^2\|\Pi_{\bW}(\mathbf{b} \bx^\intercal)\|_F^2
\big]
\label{app:eq:overlap_noise_tangent}
\end{equation}

\begin{figure}[t]
  \centering
  \includegraphics[width=0.9\linewidth]{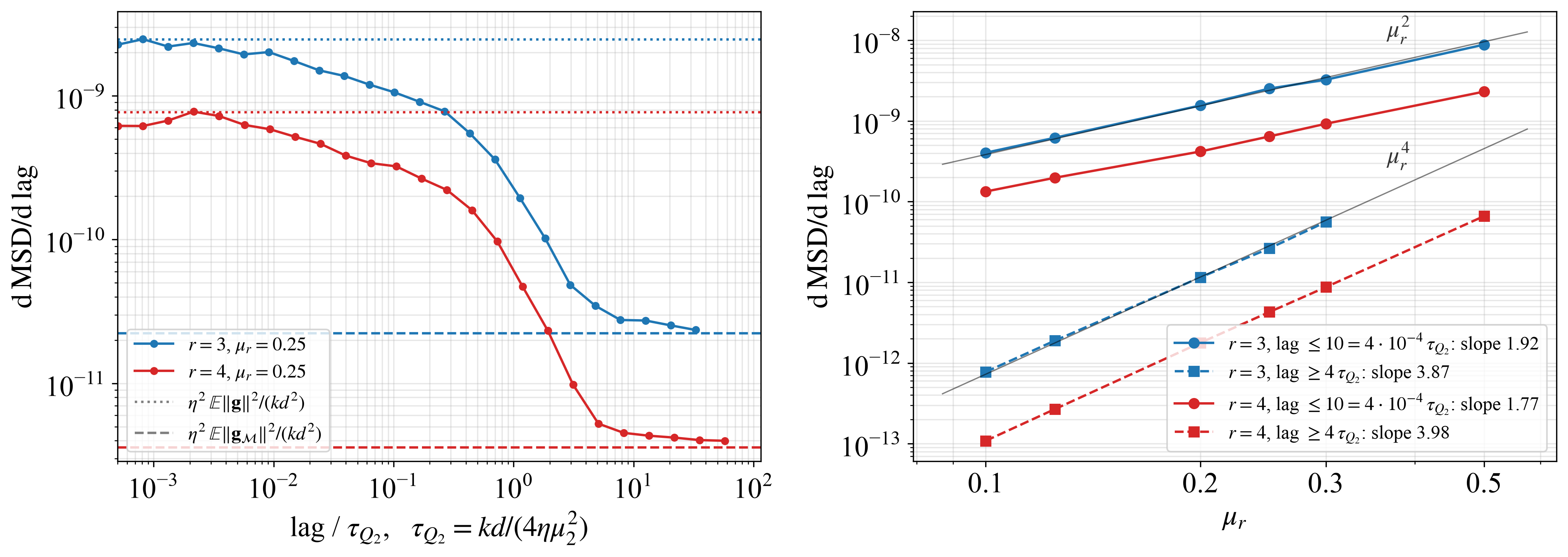}
  \caption{\label{fig:quadratic-shielding}\textbf{Experimental verification of
    quadratic shielding.} We measure the mean-square displacement
    (MSD)~\cref{app:eq:msd} of the teacher--student overlaps $\Omega_{ij}^{01}$
    on the Gaussian plateau. We use the effective activation
    $\sigma(z)=\mathrm{He}_1(z)+\tfrac12\mathrm{He}_2(z)
    +\tfrac{\mu_r}{r!}\mathrm{He}_r(z)$, with $r=3,4$, $k=d=100$, $\eta=0.1$.
    Here, $\tau_{Q_2}=kd/(4\eta\mu_2^2)$ is the relaxation time of the $Q_2$
    dynamics, as per \cref{eq:q2_dyn}. \textbf{Left:} The MSD slope as a
    function of the lag time shows a crossover between two noise scales ($r=3,4,
    \mu_r=0.25$) \textbf{Right:} The two limiting MSD slopes as a function of
    $\mu_r$ for short and long lags compared to $\tau_{Q_2}$.}
\end{figure}

To test whether this projected noise scale describes ordinary SGD over longer
lags, we analyse the dyanmics of $n=16$ students trained on the same teacher
from the same initial weights, but using different input streams that drive the
dynamics. We measure the centered
mean-square displacement of an overlap $\omega$ on the plateau,
\begin{equation}
    \label{app:eq:msd}
    M(t)\coloneq \var_a [ \omega ^a(t_0+t)-\omega^a(t_0)],
\end{equation}
where $a$ labels replicas. We use $\tau_{Q_2}=kd/(4\eta\mu_2^2)$ as the
characteristic quadratic relaxation scale.  We see on the left of
\cref{fig:quadratic-shielding} that for short times compared to $\tau_{Q_2}$,
the slope of the MSD is indeed given by the full instantaneous noise and scales
as $\varepsilon^2$. For longer times $t > \tau_{Q_2}$ instead, quadratic
shielding kicks in and reduces the SGD noise, leaving an MSD slope proportional
to $\varepsilon^4$. Since all replicas start \emph{after} quadratic learning,
this crossover reflects the relaxation of newly injected normal fluctuations,
not further learning of the quadratic target. These observations support the
effective-confinement description of ordinary SGD over the measured plateau
window: the one-step noise remains quadratic in $\varepsilon$, while its
accumulated effect along the slow directions is quartic. We repeated the
experiment with various values of $\mu_r$, or equivalently, $\epsilon_r$, and
find that the trend persists at short and long lag times as predicted.

\section{Dynamics of the refinement regime}
\label{sec:self-consis}

\subsection{Fixed point}
We next derive the dynamics after the specialization transition has selected a permutation of the hidden neurons. After matching, the diagonal student--teacher overlaps are $O(1)$, whereas the off-diagonal overlaps remain at the Gaussian background scale,
$\Omega_{ii}^{01}=O(1)$, $\Omega_{ij}^{01}=O(k^{-1/2}), i\ne j$. Therefore, for any $p>2$,
\begin{equation*}
    Q_p^{01}
    =
    \frac{1}{k}\sum_{i=1}^k(\Omega_{ii}^{01})^p
    +
    \frac{1}{k}\sum_{i\ne j}(\Omega_{ij}^{01})^p .
\end{equation*}
The first term is $O(1)$, while the off-diagonal Gaussian contribution scales as $ O(k^{1-\lceil p/2\rceil})$, and thus can be ignored.

The emergence of matched student--teacher overlaps does not require modifying the $Q_1$ and $Q_2$ equations: the contraction calculations of \cref{Sec:contractions} already include the relevant diagonal index configurations.

To see this, consider the three-point contraction in \cref{eq:three_point}:
$$
\frac{1}{k^2}\sum_{i,j,m=1}^k
\Omega^{01}_{im}\Omega^{00}_{ij}
=
\frac{1}{d^2k^2}
\sum_{i,j,m=1}^k
\sum_{\alpha,\beta=1}^d
W_{i\alpha}^0W_{m\alpha}^1W_{i\beta}^0W^0_{j\beta}.
$$
In the original derivation, this contraction was separated according to the input-coordinate indices. The contribution with $\alpha=\beta$ gives the connected term proportional to $Q_1^{01}$, while the contribution with $\alpha\ne\beta$ factorizes at leading order into a product of two single-overlap sums, giving $Q_1^{01}Q_1^{00}$.

In the refinement regime, configurations with $i=m$ give large contributions. These configurations can themselves be divided into $i=m,\alpha=\beta$ and $i=m,\alpha\ne\beta$. Both cases are already contained in the two classes of the original contraction calculation. In particular, the order parameter $Q_1^{01}$ denotes the full first-order cross-overlap,
$Q_1^{01}   = Q_{W_1}^{01} + Q_{1,\mathrm{off}}^{01},$
and therefore includes both the matched diagonal contribution and the off-diagonal background. The same point applies to the loop contractions. For instance, consider ${k^{-2}}\sum_{i,j,m=1}^k \Omega^{01}_{jm}\Omega^{01}_{im}\Omega^{00}_{ij}$.
In the coordinate-level expansion discussed in \cref{Sec:contractions}, these configurations are further classified according to coincidences among input indices and hidden-neuron indices. However, a configuration such as $j=m, \alpha\ne\beta\ne\nu, i\ne j$ does not generate a new contribution. The diagonal contribution has already been  absorbed into conditions discussed before. Thus no additional diagonal-spike term should be added to the $Q_1$ or $Q_2$ equations unless one rewrites the dynamics explicitly in terms of residual off-diagonal fields.

We now consider the higher-order parameters, which are dominated by the matched student--teacher overlaps. Keeping the order-$\eta$ terms after the global Frobenius normalization, we obtain the dynamics of the matched overlaps:
\begin{equation}
        \Omega_{ii}^{01}(t+1)-\Omega_{ii}^{01}(t)
        =
        \frac{2\eta}{kd}
        \sum_{\ell\ge0}
        \frac{\mu_{\ell+1}^2}{\ell!}
        \Big[
            \sum_{j=1}^k
            \Big(
            (\Omega_{ij}^{10})^\ell\Omega_{ij}^{00}
            -
            (\Omega_{ij}^{11})^\ell\Omega_{ij}^{01}
            \Big)
            -
            \Omega_{ii}^{01}
            \big(
            Q_{\ell+1}^{01}-Q_{\ell+1}^{11}
            \big)
            \Big]
        +
        O(\eta^2).
    \label{eq:diag_Q1_full}
\end{equation}
The $\mu_1$ sector contributes
$Q_1^{00}-Q_1^{01}-\Omega_{ii}^{01}(Q_1^{01}-Q_1^{11}),$ while the $\mu_2$ sector contributes $\Omega_{ii}^{01}(Q_2^{11}-Q_2^{01})$ since $\sum_{j=1}^k \Omega_{ij}^{10} \Omega_{ij}^{00} \approx            \sum_{j=1}^k\Omega_{ij}^{11}\Omega_{ij}^{01} \approx(1+k/d)\Omega_{ii}^{01}$.

We now investigate the fixed point of \cref{eq:diag_Q1_full}, where the closed $Q_1$ and $Q_2$ dynamics have already reached their plateaus. There, the $\ell+1=1$ and $\ell+1=2$ sectors vanish at leading order, so the remaining drift comes only from the Hermite components with $\ell+1\ge 3$. In the refinement regime, the corresponding cross-overlaps $Q_{\ell+1}^{01}$ are dominated by the matched diagonal entries. Thus,
\begin{equation}
    Q_{\ell+1}^{01}-Q_{\ell+1}^{11}
    =
    \frac{1}{k}\sum_{s=1}^k
    (\Omega_{ss}^{01})^{\ell+1}
    -
    1.
    \label{eq:global_diag_closure}
\end{equation}
The same diagonal closure applies to the local term in \cref{eq:diag_Q1_full}. For $\ell\ge2$,
\begin{equation}
    \begin{aligned}
        \sum_{j=1}^k
        \Big[
            (\Omega_{ij}^{10})^\ell\Omega_{ij}^{00}
            -
            (\Omega_{ij}^{11})^\ell\Omega_{ij}^{01}
        \Big]
        & =
        (\Omega_{ii}^{01})^\ell-\Omega_{ii}^{01},
    \end{aligned}
    \label{eq:local_diag_closure}
\end{equation}
where the leading contribution comes from the term $j=i$.
Substituting \cref{eq:local_diag_closure} and~\cref{eq:global_diag_closure} into \cref{eq:diag_Q1_full} gives the closed diagonal dynamics
\begin{equation}
        \Omega_{ii}^{01}(t+1)-\Omega_{ii}^{01}(t) \label{eq:diag_Q1_closed}
        =
        \frac{2\eta}{kd}
        \sum_{\ell\ge2}
        \frac{\mu_{\ell+1}^2}{\ell!}
        \Big[
            (\Omega_{ii}^{01})^\ell
            -
            \Omega_{ii}^{01}
            \frac{1}{k}\sum_{s=1}^k
            (\Omega_{ss}^{01})^{\ell+1}
        \Big]+
        O(\eta^2).
\end{equation}
At a stationary point of the leading diagonal flow, \cref{eq:diag_Q1_closed} yields the self-consistency equation
\begin{equation*}
    \sum_{\ell\ge2}
    \frac{\mu_{\ell+1}^2}{\ell!}
    \Big[
        (\Omega_{ii}^{01})^\ell
        -
        \Omega_{ii}^{01}
        \frac{1}{k}
        \sum_{s=1}^k
        (\Omega_{ss}^{01})^{\ell+1}
    \Big]
    =
    0,
\end{equation*}
for $i=1,\dots,k$. A sufficient construction is to require the contribution from each Hermite sector to vanish separately: for every $\ell\ge2$ with $\mu_{\ell+1}\ne0$, nonzero matched overlaps must satisfy $(\Omega_{ii}^{01})^{\ell-1}=k^{-1}\sum_{s=1}^k(\Omega_{ss}^{01})^{\ell+1}$ for $i=1,\dots,k$. Since the right-hand side is independent of $i$, all nonzero matched overlaps have the same allowed magnitude. On the fully positive aligned branch, this identifies the homogeneous stationary configuration $\Omega_{ii}^{01}=1$.

\begin{figure}[t]
  \centering
  \includegraphics[width=0.7\linewidth]{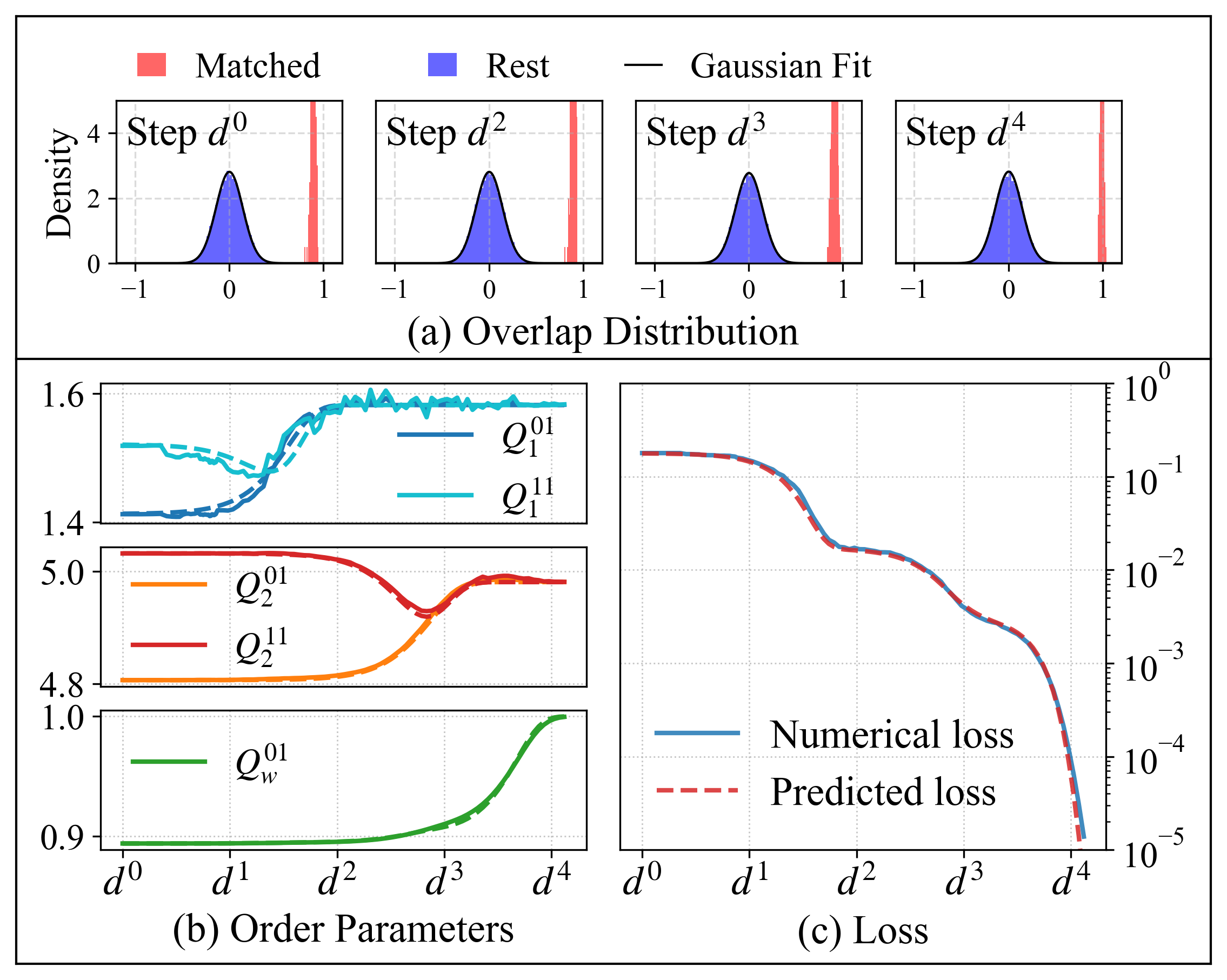}
  \caption{\label{fig:informed-init} Matched-initialization experiment and
     full-trajectory macroscopic prediction.  The student is initialized with
     partial teacher information according to \cref{eq:init}, with $s=0.5$, ELU
     activation, $d=50$, $k=200$, and learning rate $\eta=0.1$.  (a)
     Student--teacher overlap distributions at different training times, plotted
     as in \cref{fig:op-histogram}. The matched entries form a diagonal spike,
     while the remaining overlaps are compared with a Gaussian fit.  (b) Time
     evolution of the macroscopic order parameters. Solid curves show the
     simulation results, and dashed curves show the theoretical ODE predictions.
     (c) Population loss along training.}
\end{figure}
\subsection{Informed initialization}
\label{sec:informed_init}
To extend the Gaussian-closure description across the full training trajectory, we consider a constructed matched-initialization setting in which the student weight matrix is initialized with partial information from the teacher,
\begin{equation}
    \bW(t=0) = \sqrt{kd}\frac{\bW^0 + s\bZ}{\|\bW^0 + s\bZ\|_{\rm F}},\label{eq:init}
\end{equation}
where $\bZ$ is a standard Gaussian noise matrix and $s$ is a control parameter that tunes the distance from the teacher initialization. Under this initialization, the student weights already contain a planted teacher-aligned component. Consequently, the overlap distribution is well described by a spike-plus-Gaussian ansatz throughout training: the matched diagonal component evolves as a macroscopic spike, while the remaining off-diagonal overlaps stay close to a Gaussian background. The low-order overlaps $Q_1^{01}$ and $Q_2^{01}$ may already be large at initialization, but they have not necessarily reached their fully learned plateaus, so their dynamics must also be tracked. Under the spike-plus-Gaussian ansatz, the matched diagonal overlaps are assumed to concentrate around a common value, $
    \Omega_{ii}^{01}\approx Q_{W_1}^{01},i=1,\ldots,k .$ Consequently, the higher-order cross-overlaps are dominated by the diagonal spike and satisfy, for $p\ge3$,
$Q_p^{01} \approx (Q_{W_1}^{01})^p, Q_p^{11}\approx 1 $. Thus, within this ansatz, the training trajectory can be described by five macroscopic order parameters:
$Q_1^{01},Q_1^{11}, Q_2^{01},Q_2^{11}, Q_{W_1}^{01}$. The first four are governed by the closed $Q_1$ and $Q_2$ equations derived above. The remaining equation for $Q_{W_1}^{01}$ follows from \cref{eq:diag_Q1_full}:
\begin{align*}
Q_{W_1}^{01}(t+1)
={}&
Q_{W_1}^{01}(t)
+
\frac{2\eta}{kd}
\Big[
\mu_1^2
\Big(Q_1^{00}-Q_1^{01} - Q_{W_1}^{01}(Q_1^{01}-Q_1^{11}) \Big)
+ \mu_2^2 Q_{W_1}^{01}(Q_2^{11}-Q_2^{01})
\\
&+ \sum_{\ell\ge2}
\frac{\mu_{\ell+1}^2}{\ell!}
\Big((Q_{W_1}^{01})^\ell- (Q_{W_1}^{01})^{\ell+2} \Big) \Big].
\end{align*}

\Cref{fig:informed-init} shows the overlap distributions and the time evolution of the order parameters for $s=0.5$ and ELU activation. The matched entries form a diagonal spike while the remaining overlaps stay Gaussian, and the five-variable flow reproduces the measured order parameters; the corresponding loss is shown in panel (c).

\end{document}